\documentclass{article}
\usepackage{iclr2027_conference,times}

\usepackage{amsmath,amsfonts,bm}

\def\eqref#1{equation~\ref{#1}}

\def\1{\bm{1}}

\DeclareMathAlphabet{\mathsfit}{\encodingdefault}{\sfdefault}{m}{sl}
\SetMathAlphabet{\mathsfit}{bold}{\encodingdefault}{\sfdefault}{bx}{n}

\usepackage{hyperref}
\usepackage{url}
\usepackage{booktabs}
\usepackage{wrapfig}
\usepackage{graphicx}
\usepackage{booktabs}
\usepackage{amsmath}
\usepackage{float}
\usepackage{subcaption}
\usepackage{multirow}

\title{Suppressing the Shortcut Hidden Behind the Gains of VideoLLM Fine-Tuning}

\author{Youngwoo Shin$^{1}$ \qquad Jongsuk Kim$^{2,\dagger}$ \qquad Junmo Kim$^{1,\dagger}$ \\[2pt]
$^{1}$Korea Advanced Institute of Science and Technology (KAIST) \qquad $^{2}$PYLER \\[2pt]
\texttt{\{yshin0917, junmo.kim\}@kaist.ac.kr} \quad \texttt{js.kim@pyler.tech}
}

\newcommand{\codelink}{ Code is available at \url{https://github.com/Youngwoo-git/VETO}.}

\iclrfinalcopy
\begin{document}

\maketitle
\vspace{-0.2in}
{\renewcommand{\thefootnote}{\fnsymbol{footnote}}\footnotetext[2]{Corresponding authors.}}
\lhead{Preprint}

\providecommand{\codelink}{}
\begin{abstract}

Supervised fine-tuning (SFT) is the stage in which Video Large Language Models (VideoLLMs) advance their capability to reason over video. However, the VideoLLM also learns to answer from textual regularities of its training data, a shortcut that bypasses the video. The regularities behind the shortcut are hard to define at the data level, and the shortcut forms regardless of training epoch, data size, or backbone. The shortcut is hidden behind a rise in accuracy within the training distribution, yet it is not required for the rise and even costs SFT the gain the video offers under distribution shift. We therefore propose Video-Evidence Training via Orthogonal projection (VETO), which intervenes on the fine-tuning update where the shortcut is written rather than on the data. VETO calibrates the subspace that the text alone occupies in the activations of the base model and projects the update onto its orthogonal complement. The VideoLLM is thus left to ground its answer in the full input, with the shortcut suppressed as it forms and never defined. Across seven VideoLLMs, VETO preserves the gain of SFT on the training distribution and suppresses the shortcut. On six benchmarks under shift it gains more than SFT and answers from the full input rather than the shortcut.\codelink

\end{abstract}

\vspace{-9pt}

\section{Introduction}
\label{sec:intro}

Video Large Language Models (VideoLLMs) extend the reasoning that Large Language Models (LLMs) acquire from text to video, an input of rich, dynamic visual semantics~\citep{bai2025qwen25vltechnicalreport, bai2025qwen3vltechnicalreport, li2025llavaonevision, clark2026molmo2openweightsdata, gemmateam2026gemma4technicalreport}. Supervised fine-tuning (SFT) remains the established stage for adapting a pretrained VideoLLM to a downstream task~\citep{bai2025qwen25vltechnicalreport} and grounding that reasoning in the video. Even recent methods trained with reinforcement learning (RL) retain this stage as the basis of their improvement~\citep{wang2025videorft, wang2025video}. While SFT is thus a core element in advancing VideoLLMs, it is prone to learning superficial correlations rather than reading the modality it is given.

These correlations, which we call shortcuts~\citep{geirhos2020shortcut, dancette2021beyond}, are regularities of the training set that predict the label within the training distribution but may not hold outside it. Video QA is especially susceptible, as many questions can be answered from regularities in the text alone~\citep{yang2025vidlbeval}, and SFT readily comes to rely on them~\citep{li2023evaluating, chen2026language}. Existing approaches counter a shortcut by filtering it from the data~\citep{chen2020counterfactual, shvetsova2025unbiasing}, penalizing it in the loss~\citep{cadene2019rubi, clark2019don, niu2021counterfactual, chen2026language}, or adjusting the trained model or its decoding afterward~\citep{wang2024mdpo, yang2025vidlbeval}. Filtering has to define which regularity counts as the shortcut before it can act~\citep{dancette2021beyond}, while the regularities fine-tuning will pick up are rarely known in advance, and the loss and the trained model constrain what the model outputs rather than the update through which the shortcut is written. A remedy would instead have to act where the shortcut forms, which raises our central question: \emph{where does the shortcut form in the course of fine-tuning, and can fine-tuning be constrained there to suppress it?}

To address this question, we track how the reliance of a VideoLLM on the text changes over SFT, both within the training distribution and under distribution shift. We fine-tune on filtered video QA data where the text alone gives the base model little ground for its answer. After training, the VideoLLM answers far more questions from the text alone than the base model does, regardless of training epoch, data size, or backbone. Within the training distribution, this may seem to cost nothing, since full-input accuracy rises alongside it, as Fig.~\ref{fig:teaser}(a) shows. Under the shift, however, the gain of the fine-tuned model over the base model is limited, while its accuracy from the text alone still rises above the base model on every benchmark. In other words, the shortcut fine-tuning added persists where it no longer holds. SFT can thus appear to fine-tune the model successfully while silently teaching it to rely on spurious correlations rather than the video, leaving it vulnerable to any departure from its training distribution and diluting what makes it a VideoLLM. Filtering the data before fine-tuning does not prevent the shortcut, and the rise in accuracy hides it afterward, so the intervention must come during fine-tuning, where the model can be made to ground its answer in the full input rather than the shortcut alone.

\IfFileExists{figs/fig_problem.pdf}{%
\begin{figure}[t]
  \centering
  \includegraphics[width=0.96\linewidth]{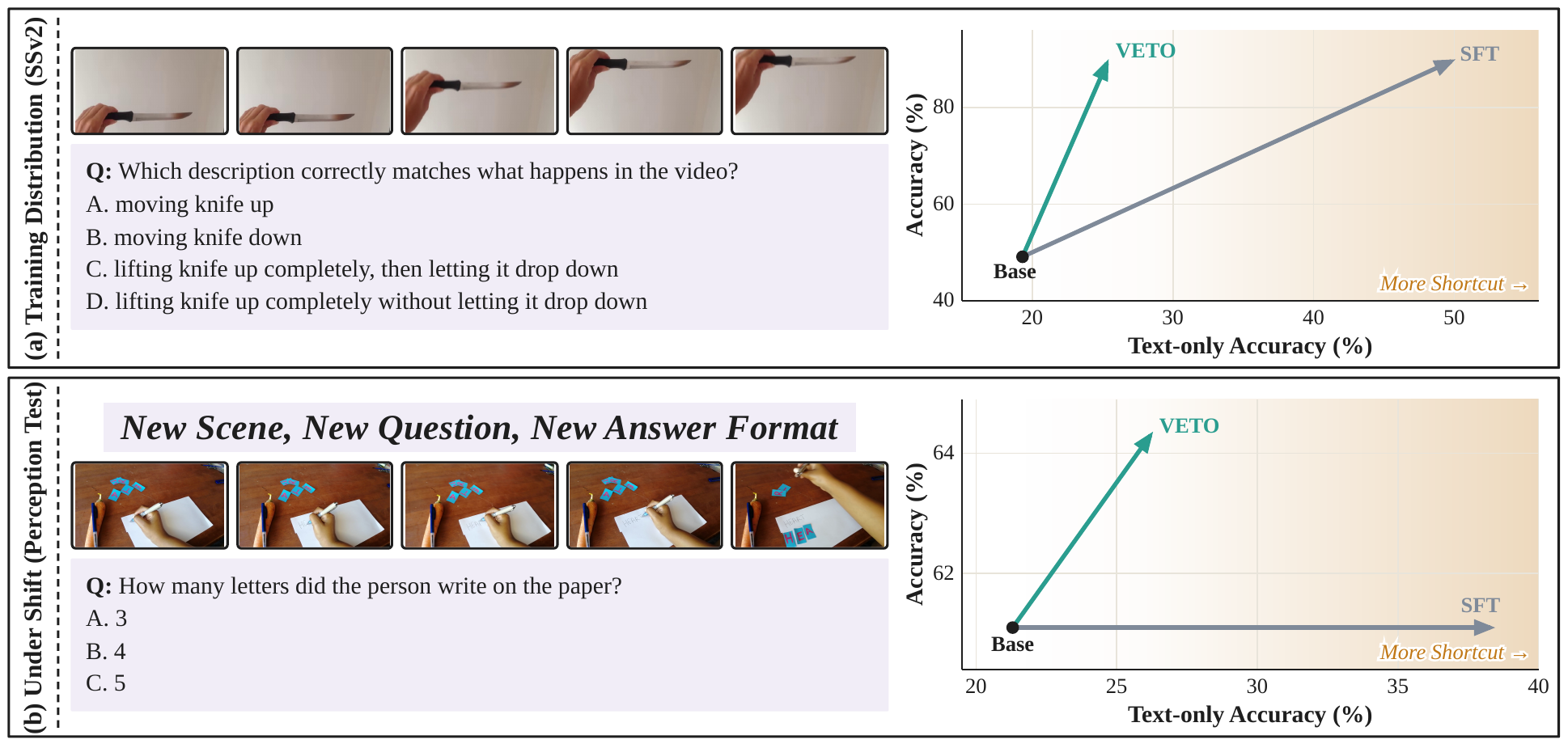}
  \caption{\textbf{SFT builds a shortcut, and VETO suppresses it.} Accuracy of Qwen2.5-VL-7B against its text-only accuracy, where a rise indicates the shortcut. (a) Within the training distribution (SSv2) SFT and VETO reach a similar gain in accuracy, but the shortcut grows much more under SFT. (b) Under the shift (Perception Test) VETO gains accuracy whereas SFT gains only the shortcut.}
  \label{fig:teaser}
  \vspace{-10pt}
\end{figure}

}{}

This motivates our method, \textbf{Video-Evidence Training via Orthogonal projection} (VETO), which keeps the fine-tuning update out of the directions where the text alone is read. VETO first calibrates the subspace carrying the shortcut from the activations produced by the text-only input at every layer, once per backbone. It then projects the LoRA update onto the orthogonal complement of that subspace during fine-tuning, so that the update cannot enter the directions in which the shortcut would be learned. As VETO constrains only the update, it leaves the prior the base model already carries untouched and allows the model to adapt to the video rather than to the shortcut.

We evaluate VETO on the training distribution across seven VideoLLMs from five language models and on six unseen benchmarks. On the training distribution, VETO reaches nearly the same gain as SFT on every backbone, while the rise in text-only accuracy under SFT is largely suppressed. Under the shift VETO gains more over the base model than SFT, as Fig.~\ref{fig:teaser}(b) illustrates, and its reliance on the text rises far less than under SFT. VETO requires no definition of which regularity counts as the shortcut, no change to the data, loss, or schedule, and no cost beyond a projection per step. Across experiments, the shortcut emerges not as an inevitable prior that fine-tuning must keep for its gain but as what stands between a VideoLLM and the video it is given.

\section{Shortcut Formation under Fine-Tuning}
\label{sec:problem}

In this section we ask how the reliance of a VideoLLM on the text changes over fine-tuning under the standard recipe for adapting it to a target video QA task, SFT with LoRA~\citep{hu2021lora}. The base model is Qwen2.5-VL-7B~\citep{bai2025qwen25vltechnicalreport} and the target dataset is Something-Something~v2 (SSv2)~\citep{goyal2017something}, whose answers come from a fixed set of action templates. Under distribution shift we evaluate on Perception Test~\citep{patraucean2023perception}, whose videos, questions, and options differ from SSv2 and whose options are written freely rather than from templates.

\subsection{Blind Filtering and the Shortcut}
\label{sec:setup}

\paragraph{Dataset preparation} starts from the action template SSv2 assigns to every clip. We group the templates into families, where the templates of a family share the same objects and scene and differ only in what happens over time. Examples are direction (\emph{moving X left} vs.\ \emph{moving X right}) and state transition (\emph{covering X} vs.\ \emph{uncovering X}). The templates of a family are thus hard negatives of one another, and no other template can serve as one. Each clip becomes one \emph{item}, a multiple-choice question in which the template of the clip is the correct option and the distractors are other templates rendered with the objects of the clip, with the families listed in Appx.~\ref{app:families}.

\paragraph{Blind filtering} removes the items that the base model already answers from the text alone. The \emph{full} input pairs the video with the question and options, and the \emph{text-only} input is the full input without the video. We query the base model three times per item under the text-only input with the option order shuffled each time and remove the item when it answers correctly in all three, so that a lucky guess does not count. The filter applies to the training and evaluation splits alike, and on the evaluation split the base model reaches $19.3\%$ text-only accuracy, below chance, as Appx.~\ref{app:filtering} details. The \emph{shortcut} is therefore any rise in text-only accuracy over fine-tuning, a route to the answer that runs through the text alone and bypasses the video.

\begin{figure}[H]
  \centering
  \begin{minipage}[t]{0.64\linewidth}
    \includegraphics[width=\linewidth]{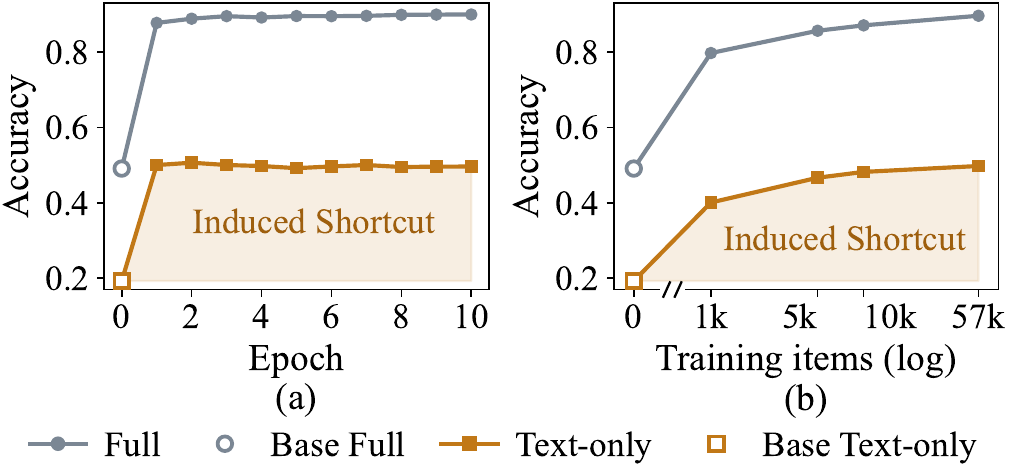}
    \captionof{figure}{\textbf{The shortcut forms regardless of training epoch and data size.} (a)~The shortcut is present from the first epoch and persists. (b)~The shortcut is substantial at every training set size, already at 1k items. Mean over three seeds.}
    \label{fig:epoch}
  \end{minipage}%
  \hfill
  \begin{minipage}[t]{0.31\linewidth}
    \includegraphics[width=\linewidth]{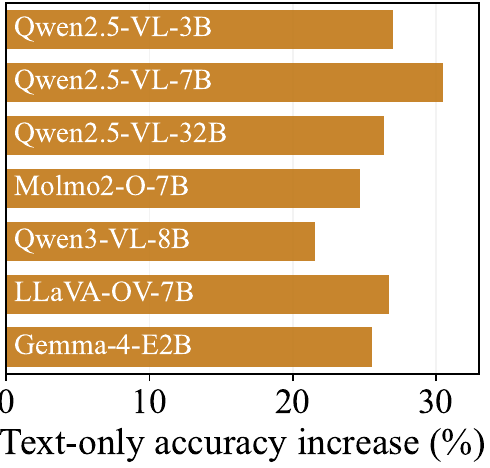}
    \captionof{figure}{\textbf{The shortcut forms regardless of backbone.} Every VideoLLM gains large text-only accuracy after SFT.}
    \label{fig:family}
  \end{minipage}
  \vspace{-4pt}
\end{figure}

\subsection{Blind Filtering Does Not Prevent Shortcut Formation}
\label{sec:bomb}

Blind filtering removes every training item that the base model solves from the text alone, so SFT would seem to have nothing to learn from the text. Yet as Fig.~\ref{fig:epoch}(a) shows, the text-only accuracy rises together with the accuracy on the full input over the course of SFT, and the model ends up solving a substantial fraction of the evaluation set without the video. SFT thus builds a route through the text that the base model does not have.

\paragraph{The shortcut forms regardless of training epoch.} Fig.~\ref{fig:epoch}(a) tracks text-only accuracy over epochs, and it is far above chance from the first epoch. The shortcut is thus not overfitting late in training but part of what SFT learns from the start, and early stopping cannot avoid it.

\paragraph{The shortcut forms regardless of data size.} Fig.~\ref{fig:epoch}(b) plots the text-only accuracy against the training set size from 1k to all 57k items, and it is far above chance already at 1k and grows with the data as the accuracy does. The shortcut is thus not the noise of a small sample that more data would average out, but a regularity of the distribution that any sample of it carries.

\paragraph{The shortcut forms regardless of backbone.} Fig.~\ref{fig:family} gives the text-only gain for seven VideoLLMs from five language models and sizes up to 32B, and each answers at least a fifth more of the evaluation set from the text alone than its base model does. The shortcut is thus not particular to one language model or scale but what SFT does with the data.

These three observations suggest that what the shortcut rests on is present regardless of the training schedule, the data scale, or the backbone. Blind filtering leaves only the items that the base model does not solve from the text alone, but that does not mean their text carries no information about the answer. The options each item receives in the dataset preparation are distinct sentences, and any regularity between their wording and the correct answer is there for SFT to pick up even where the base model does not read it. The regularity thus persists in the data under every variation, and it becomes the shortcut once SFT learns to read it.

\subsection{Why the Shortcut Matters}
\label{sec:why}

The purpose of SFT is to fit the target distribution for downstream tasks, and when that distribution carries regularity between the question and the answers, learning it can form a shortcut. From a Bayesian perspective this may not be a defect. A model that estimates a prior from training data and combines it with the visual evidence at inference does what the objective asks, and a rise in accuracy is consistent with that. Whether the shortcut is a problem therefore turns on two properties of what has been learned: what the shortcut reads and whether the visual evidence still enters the decision.

\begin{wraptable}[11]{r}{0.35\linewidth}
  \vspace{-\baselineskip}
  \centering
  \small
  \caption{Accuracy with the full input and from the text alone, on SSv2 and Perception Test.}
  \label{tab:e8}
  \setlength{\tabcolsep}{4pt}
  \begin{tabular}{@{}llcc@{}}
    \toprule
    & & Full & Text-only \\
    \midrule
    \multirow{2}{*}{SSv2}  & Base & 49.1\% & 19.3\% \\
                           & SFT  & 89.6\% & 49.8\% \\
    \midrule
    \multirow{2}{*}{\begin{tabular}[c]{@{}l@{}}Perception\\Test\end{tabular}} & Base & 61.1\% & 21.3\% \\
                                & SFT  & 61.1\% & 38.3\% \\
    \bottomrule
  \end{tabular}
\end{wraptable}
What the shortcut reads decides where it remains valid. The language prior the base model already carries reflects which actions tend to occur with which objects, and it holds where those facts hold. 
The shortcut is instead a regularity of the training set, and such a regularity may not hold once the questions and options are written differently. Tab.~\ref{tab:e8} tests this prediction on Perception Test, our evaluation set under shift, filtered as SSv2 is. On SSv2 the gap between the accuracy with the video and the accuracy from the text alone widens after fine-tuning, so the model gains more from the video than before and the shortcut sits alongside the reading of the video rather than replacing it. On Perception Test the same model reaches the accuracy of the base model it is fine-tuned from, while its accuracy from the text alone rises by $17.0$ points, so the gap narrows and the shortcut has displaced much of what it drew from the video. What fine-tuning adds still raises the text-only accuracy on a format it was never trained on, yet there it replaces part of the video reading and adds nothing to accuracy. The shortcut is therefore a problem rather than a prior the model needs. What it reads holds only within the training distribution, and under the shift it displaces the visual evidence instead of adding to it. The remaining question is where to intervene.

The training data are already blind-filtered, yet the shortcut forms through them. Curating them further would require naming the regularity, which Sec.~\ref{sec:bomb} does not. Once training ends, the shortcut is hard to detect, which leaves little room for a remedy before or after training. Intervening on the training schedule is no easier, since the shortcut is in place regardless of epoch, data size, or backbone. What remains is training itself, and a constraint at that stage would not teach the model more than fine-tuning already does, but it can suppress the shortcut as it forms.

\section{Video-Evidence Training via Orthogonal Projection}
\label{sec:veto}

\begin{figure}[t]
  \centering
  \includegraphics[width=0.91\linewidth]{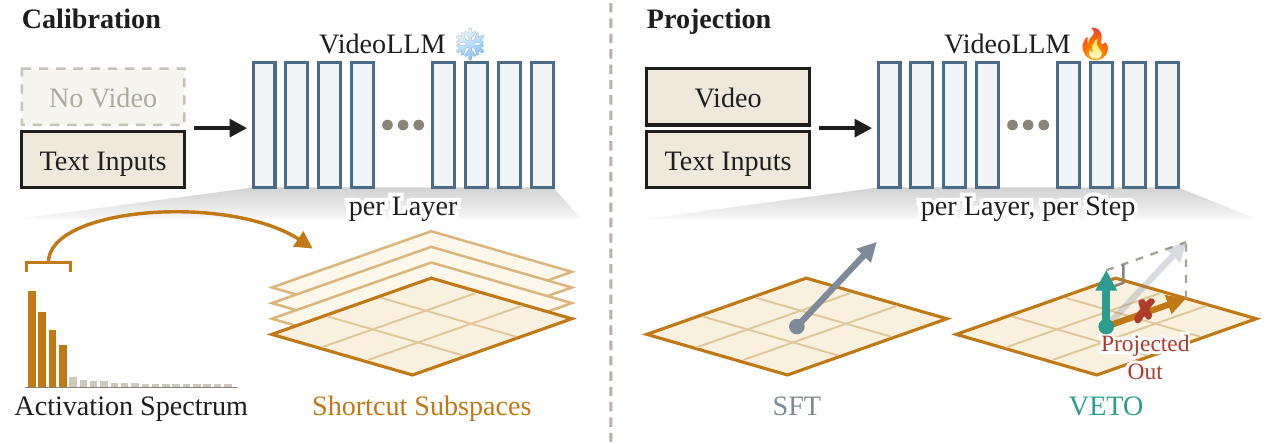}
  \caption{\textbf{Pipeline of VETO.} Calibration, once per VideoLLM, collects the activations of the text-only input at every layer and takes the leading directions of their spectrum as the shortcut subspace. During fine-tuning, the LoRA update is projected off that subspace after each optimizer step, so that the part along the shortcut directions is removed and the rest is kept, whereas SFT keeps both.}
  \label{fig:method}
  \vspace{-7pt}
\end{figure}

In this section we propose Video-Evidence Training via Orthogonal projection (VETO), which constrains the fine-tuning update to the orthogonal complement of the subspace the text alone occupies, as Fig.~\ref{fig:method} outlines. We first identify the input directions that carry the shortcut (Sec.~\ref{sec:subspace}) and then keep the update out of those directions (Sec.~\ref{sec:projection}).

\subsection{Locating the Shortcut Subspace}
\label{sec:subspace}

We identify the subspace carrying shortcut information by collecting activations under text-only input, where the model receives the question and its options but no video. Let $\mathcal{D}_{\text{cal}} = \{q_i\}$ be a calibration set drawn from the same pool as the training data and spread evenly across the families. It is held out of training and used without labels, as only its activations under the text-only input enter the calibration. For each linear module $m$ in layer $l$ with weight $W^{l,m} \in \mathbb{R}^{d_{\text{out}} \times d_{\text{in}}}$, we collect the input activations, and for brevity we drop the module index $m$ below.

\paragraph{Text-only activations} are collected by running a forward pass for each calibration item $q_i$ under the text-only input and taking the input activations at every position. Concatenated over all items, they form the columns of $H_{\text{text}}^l \in \mathbb{R}^{d_{\text{in}} \times n_{\text{text}}}$, where $n_{\text{text}}$ is the total number of positions over the calibration set rather than the number of items.

\paragraph{Shortcut subspace} is defined from the second-moment matrix of the text-only activations,

\vspace{-9pt}
\begin{equation}
    C^l \;=\; \frac{1}{n_{\text{text}}} H_{\text{text}}^l (H_{\text{text}}^l)^\top .
    \label{eq:cov}
\end{equation}
\vspace{-9pt}

The matrix is left uncentered so that the mean direction of the text-only activations stays inside the subspace, since the text alone occupies that direction as much as any other. We eigendecompose $C^l = U^l \Lambda^l (U^l)^\top$ with eigenvalues in descending order, and retain the smallest rank $r^l$ whose leading eigenvalues sum to at least a fraction $\rho$ of the trace,

\vspace{-9pt}
\begin{equation}
    r^l \;=\; \min \Big\{ r : \textstyle\sum_{j \leq r} \lambda_j^l \,\geq\, \rho \cdot \mathrm{tr}(C^l) \Big\},
    \label{eq:rank}
\end{equation}
\vspace{-9pt}

forming $U_1^l \in \mathbb{R}^{d_{\text{in}} \times r^l}$, whose columns span the shortcut subspace at layer $l$.

The text-only activations of items outside the calibration set fall inside $U_1^l$ to nearly the fraction $\rho$ imposed on the calibration items, and independent draws of the set span nearly the same subspace. The subspace is thus a property of the backbone rather than of the particular items it is estimated from, and a few hundred items suffice to estimate it, as Appx.~\ref{app:calib} shows. The one quantity $\rho$ sets is the retained rank $r^l$ and thereby how much of the input space the update cannot enter.

\subsection{Projecting Out the Shortcut}
\label{sec:projection}

Given the shortcut subspace identified, we keep the fine-tuning update $\Delta W^l$ out of it by projecting the update onto the orthogonal complement of the subspace, $\Delta W^l \leftarrow \Delta W^l P^l$, with the projector

\vspace{-9pt}
\begin{equation}
    P^l \;=\; I - U_1^l (U_1^l)^\top .
\end{equation}
\vspace{-9pt}

We instantiate the update with LoRA~\citep{hu2021lora}, which parameterizes it as a low-rank product $\Delta W^l = B^l A^l$, so that the projection acts on the input factor. At each optimizer step, the accumulated update is projected on its input side as $\Delta W^l P^l$ by right-multiplying the down-projection factor, $A^l \leftarrow A^l P^l$. The optimizer states are left as they are, since the projection is idempotent and returns the component of $A^l$ inside the subspace to zero, which Appx.~\ref{app:impl} verifies over training. This yields the invariance

\vspace{-9pt}
\begin{equation}
    \big(W^l + \Delta W^l P^l\big)\,\mathbf{u} \;=\; W^l \mathbf{u}
    \qquad \forall\, \mathbf{u} \in \mathrm{span}(U_1^l),
    \label{eq:invariance}
\end{equation}
\vspace{-9pt}

which guarantees that the response of the fine-tuned model to any input direction inside the shortcut subspace remains that of the base model.

Two properties follow from Eq.~\ref{eq:invariance}. First, the constraint is preventive rather than corrective, since the shortcut directions are unreachable from the first gradient step, so no pathway forms along them. Second, the constraint acts on the update alone. What the base model already reads from the text as a prior or a valid deduction stays as it is by construction, and only the text-only pathway that fine-tuning would add is suppressed. The success criterion is therefore the removal of the gap induced by fine-tuning, $\text{acc}_{\text{text-only}}^{\text{SFT}} - \text{acc}_{\text{text-only}}^{\text{base}}$, not the base prior itself.

Both properties concern the directions inside the span of $U_1^l$. Within the span the removal is exact, as $\Delta W^l P^l U_1^l$ is identically zero, and in practice the ratio $\|\Delta W^l U_1^l\|_F / \|\Delta W^l\|_F$ stays at numerical zero in every run. The span may not cover the whole shortcut, since the subspace is estimated from calibration activations, and any component outside it remains open to the update. The constraint also says nothing about where the displaced learning settles, and within the span it keeps out every new use of the text, the valid ones along with the shortcut. What VETO removes, where fine-tuning routes it, and whether the gain of fine-tuning survives the constraint are the questions for experiments.

\section{Experiments}
\label{sec:experiments}

\begin{figure}[t]
  \centering
  \includegraphics[width=0.98\linewidth]{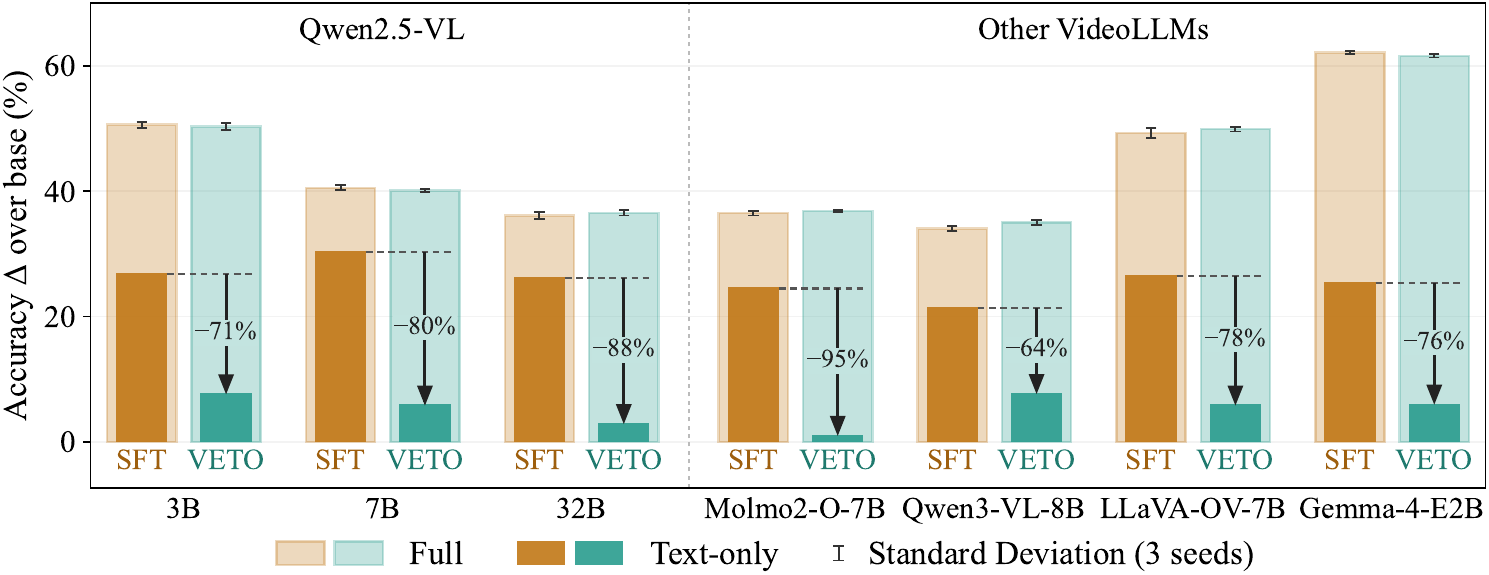}
  \caption{\textbf{VETO removes a substantial part of the shortcut across VideoLLMs.} The full-input gain is nearly the same as under SFT, whereas the text-only gain is suppressed as intended.}
  \label{fig:veto}
  \vspace{-10pt}
\end{figure}

\subsection{Setup}
\label{sec:expsetup}

\paragraph{Fine-tuning} baseline is standard SFT with rank-16 LoRA on the 57k items of the blind-filtered SSv2 training set of Sec.~\ref{sec:setup}, with 16 frames per video, for up to 10 epochs at learning rate $10^{-4}$. We select checkpoints from a held-out subset of the evaluation split that no reported accuracy uses, the same for every run. VETO trains under the identical recipe with the projection of Sec.~\ref{sec:projection} at $\rho = 0.95$. Every run uses a single NVIDIA B200 GPU, and Appx.~\ref{app:impl} lists the remaining hyperparameters, the baseline implementations, and the seed statistics.

\paragraph{Backbones} span seven VideoLLMs from five language models, Qwen2.5-VL~\citep{bai2025qwen25vltechnicalreport} at 3B, 7B, and 32B, Qwen3-VL-8B~\citep{bai2025qwen3vltechnicalreport}, LLaVA-OneVision-7B~\citep{li2025llavaonevision}, Molmo2-O-7B~\citep{clark2026molmo2openweightsdata}, and Gemma4-E2B~\citep{gemmateam2026gemma4technicalreport}, each fine-tuned with the recipe above and calibrated against its own activations.

\paragraph{Evaluation} covers the SSv2 evaluation split and six benchmarks the fine-tuning never sees, Perception Test~\citep{patraucean2023perception}, Vinoground~\citep{zhang2024vinoground}, AoTBench~\citep{xue2025seeing}, MVBench~\citep{li2024mvbench}, TVBench~\citep{cores2024lost}, and TempCompass~\citep{liu2024tempcompass}. We blind-filter each set as in Sec.~\ref{sec:setup}, and accuracy uses the highest-logit option letter as the answer. Appx.~\ref{app:filtering} gives the counts.

\subsection{Fine-Tuning Gains Without the Shortcut}
\label{sec:main}

We measure accuracy changes relative to the base model across model and backbone scales. Fig.~\ref{fig:veto} reports the gain with the full input and from the text alone, and the shortcut reduction $1 - \Delta_{\text{t}}^{\text{VETO}} / \Delta_{\text{t}}^{\text{SFT}}$, with $\Delta_{\text{t}}$ the text-only gain over the base model. On every VideoLLM, VETO suppresses the shortcut formation, while SFT and VETO gain alike with the full input. SFT raises text-only accuracy along with full-input accuracy, whereas VETO raises it far less. The shortcut that rests on the regularities of the training set may offer the VideoLLM a ground for its answers on the training distribution, but it is not indispensable. The VideoLLM trained with VETO reaches nearly the same gain without it, so VETO suppresses no prior that accuracy requires.

\begin{table}[H]
  \centering
  \caption{Accuracy (Acc.) and text-only gain over the base model (Text $\Delta$) of Qwen2.5-VL-7B on the blind-filtered subset of Perception Test, by area. Best in \textbf{bold}, second best \underline{underlined}.
  }
  \label{tab:shift}
  \footnotesize
  \setlength{\tabcolsep}{2pt}
  \begin{tabular}{@{}l ll ll ll ll ll@{}}
    \toprule
    & \multicolumn{2}{c}{Physics} & \multicolumn{2}{c}{Semantics} & \multicolumn{2}{c}{Abstraction} & \multicolumn{2}{c}{Memory} & \multicolumn{2}{c}{All} \\
    \cmidrule(lr){2-3}\cmidrule(lr){4-5}\cmidrule(lr){6-7}\cmidrule(lr){8-9}\cmidrule(lr){10-11}
    & Acc.$\uparrow$ & Text $\Delta\downarrow$ & Acc.$\uparrow$ & Text $\Delta\downarrow$ & Acc.$\uparrow$ & Text $\Delta\downarrow$ & Acc.$\uparrow$ & Text $\Delta\downarrow$ & Acc.$\uparrow$ & Text $\Delta\downarrow$ \\
    \midrule
    Base & 56.9 & -- & \underline{78.0} & -- & \underline{52.4} & -- & 41.5 & -- & \underline{61.1} & -- \\
    SFT & \underline{59.0}{\tiny$\pm$2.0} & +17.4{\tiny$\pm$1.8} & 75.0{\tiny$\pm$1.0} & +17.7{\tiny$\pm$2.9} & 52.2{\tiny$\pm$0.6} & +17.5{\tiny$\pm$2.0} & \textbf{45.0}{\tiny$\pm$1.5} & +6.2{\tiny$\pm$4.2} & \underline{61.1}{\tiny$\pm$0.7} & +17.0{\tiny$\pm$1.1} \\
    \textbf{VETO} & \textbf{63.0}{\tiny$\pm$1.1} & \textbf{+6.7}{\tiny$\pm$3.8} & \textbf{78.5}{\tiny$\pm$0.6} & \textbf{+4.0}{\tiny$\pm$1.6} & \textbf{54.3}{\tiny$\pm$1.2} & \textbf{+3.9}{\tiny$\pm$0.7} & \textbf{45.0}{\tiny$\pm$0.7} & \textbf{+1.2}{\tiny$\pm$2.8} & \textbf{64.3}{\tiny$\pm$0.7} & \textbf{+4.9}{\tiny$\pm$2.2} \\
    \bottomrule
  \end{tabular}
  \vspace{-10pt}
\end{table}

\begin{table}[H]
  \centering
  \caption{Accuracy (Acc.) and text-only gain over the base model (Text $\Delta$) of Qwen2.5-VL-7B on the blind-filtered subsets of five further benchmarks, as in Tab.~\ref{tab:shift}.
  }
  \label{tab:shift_more}
  \footnotesize
  \setlength{\tabcolsep}{2pt}
  \begin{tabular}{@{}l ll ll ll ll ll@{}}
    \toprule
    & \multicolumn{2}{c}{Vinoground} & \multicolumn{2}{c}{AoTBench} & \multicolumn{2}{c}{MVBench} & \multicolumn{2}{c}{TVBench} & \multicolumn{2}{c}{TempCompass} \\
    \cmidrule(lr){2-3}\cmidrule(lr){4-5}\cmidrule(lr){6-7}\cmidrule(lr){8-9}\cmidrule(lr){10-11}
    & Acc.$\uparrow$ & Text $\Delta\downarrow$ & Acc.$\uparrow$ & Text $\Delta\downarrow$ & Acc.$\uparrow$ & Text $\Delta\downarrow$ & Acc.$\uparrow$ & Text $\Delta\downarrow$ & Acc.$\uparrow$ & Text $\Delta\downarrow$ \\
    \midrule
    Base & 55.9 & -- & 41.3 & -- & \textbf{58.0} & -- & 42.4 & -- & 64.0 & -- \\
    SFT & \underline{63.5}{\tiny$\pm$0.4} & +13.5{\tiny$\pm$1.7} & \underline{53.1}{\tiny$\pm$0.2} & +10.9{\tiny$\pm$1.3} & 56.5{\tiny$\pm$0.8} & +7.8{\tiny$\pm$0.5} & \underline{44.3}{\tiny$\pm$0.8} & +7.6{\tiny$\pm$0.4} & \underline{65.8}{\tiny$\pm$0.6} & +4.0{\tiny$\pm$0.6} \\
    \textbf{VETO} & \textbf{66.9}{\tiny$\pm$2.0} & \textbf{+7.7}{\tiny$\pm$0.8} & \textbf{55.2}{\tiny$\pm$1.3} & \textbf{+5.6}{\tiny$\pm$2.5} & \textbf{58.0}{\tiny$\pm$1.0} & \textbf{+3.5}{\tiny$\pm$0.3} & \textbf{46.4}{\tiny$\pm$0.4} & \textbf{+3.0}{\tiny$\pm$1.5} & \textbf{67.6}{\tiny$\pm$0.3} & \textbf{+2.4}{\tiny$\pm$0.7} \\
    \bottomrule
  \end{tabular}
  \vspace{-5pt}
\end{table}

\subsection{Shortcut under Distribution Shift}
\label{sec:shift}

Tab.~\ref{tab:shift} reports the accuracy on Perception Test, whose videos, questions, and options all differ from SSv2. SFT raises text-only accuracy by $17.0$ points overall and more than VETO in every area, so the shortcut of SFT carries over to this benchmark. On the full input, however, SFT is no better than the base model, whereas VETO improves over the base model while raising text-only accuracy much less. On this benchmark SFT keeps the shortcut but loses its gain, while VETO suppresses the shortcut and keeps its gain. Tab.~\ref{tab:shift_more} extends the comparison to five further unseen benchmarks, and VETO exceeds SFT on every one of them. VETO also brings far less of the shortcut into every benchmark than SFT, so the suppression carries over as well. Suppressing the shortcut is thus what makes the VideoLLM gain more than SFT under the shift.

\subsection{Where the VideoLLM Grounds Its Reasoning}
\label{sec:redistribution}

As VETO suppresses the shortcut while largely preserving the accuracy of the VideoLLM, a pathway other than the shortcut has to carry that accuracy. This leads us to investigate where the VideoLLM grounds its reasoning on Qwen2.5-VL-7B throughout this section. To this end, we compare SFT and VETO on the full input and the text alone using two probes.

\begin{wrapfigure}[11]{l}{0.39\linewidth}
  \vspace{-\baselineskip}
  \centering
  \includegraphics[width=\linewidth]{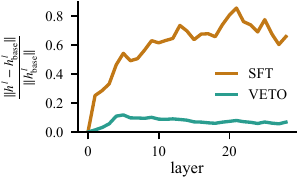}
  \setlength{\abovecaptionskip}{-8pt}
  \caption{Relative deviation from the base model per layer, text-only input.}
  \label{fig:sec44a}
\end{wrapfigure}
The first probe asks where fine-tuning changes the computation on text. Under the text-only input we measure how far fine-tuning has moved the representation at each layer. The measure is the norm of the difference between the hidden states of the last token in the fine-tuned and the base model, relative to the norm in the base model. As presented in Fig.~\ref{fig:sec44a}, SFT moves the representation further with depth, so the change that fine-tuning makes to the computation on text accumulates over the layers. VETO keeps it close to the base model at every layer. The invariance of Eq.~\ref{eq:invariance} thus carries through the network.

\begin{wraptable}[9]{r}{0.45\linewidth}
  \centering
  \footnotesize
  \setlength{\tabcolsep}{4pt}
  \caption{Entropy over the options as \% of its maximum. Lower \% for more confidence.}
  \label{tab:sec44ent}
  \vspace{-5pt}
  \begin{tabular}{@{}llccc@{}}
    \toprule
    & & Text-only & Full & Video $\Delta$ \\
    \midrule
    \multirow{2}{*}{SSv2} & SFT  & 16\% & 2\% & 14\% \\
                          & \textbf{VETO} & 68\% & 2\% & 66\% \\
    \midrule
    \multirow{2}{*}{\begin{tabular}[c]{@{}l@{}}Perception\\Test\end{tabular}} & SFT  & 33\% & 14\% & 19\% \\
                                & \textbf{VETO} & 81\% & 18\% & 63\% \\
    \bottomrule
  \end{tabular}
\end{wraptable}
The second probe asks how confident the VideoLLM is in its answer, so we take the entropy over the options as a share of its maximum. The entropy is maximal when the options are equally likely and zero when one option takes all the probability, so low entropy means the VideoLLM has settled on an answer. Tab.~\ref{tab:sec44ent} reports the entropy under the text-only and full input. Their difference is the part of the uncertainty that the video resolves and appears as Video $\Delta$. SFT already settles on an answer without the video, so little is left for the video to resolve. VETO stays unsettled without the video, and the video resolves most of that uncertainty. The confidence therefore comes from the video under VETO and from the text under SFT.

The shortcut could be taken for a prior the VideoLLM needs to reason since SFT raises the accuracy on the training distribution, but the two probes say otherwise. The first shows that computation on text alone stays close to the base model at every layer, while the VideoLLM reaches accuracy with the video, so the accuracy does not rest on the prior. The second shows that the confidence SFT gains without the video is not a prior awaiting evidence but an answer already settled, and SFT carries this settled answer under the shift, where the regularity behind it no longer holds. With VETO, the confidence appears only once the video is present, so the accuracy the VideoLLM reaches with the video is carried by its reading of the video, on the training distribution and under the shift.

\subsection{Other Interventions and Ablations}
\label{sec:ablation}

\begin{table}[H]
  \footnotesize
  \setlength{\tabcolsep}{2.5pt}
  \setlength{\aboverulesep}{0.25ex}\setlength{\belowrulesep}{0.65ex}
  \begin{minipage}[t]{0.57\linewidth}
    \centering\footnotesize
    \captionof{table}{Alternatives to VETO. Text $\Delta$ is the text-only gain over the base model, lower for more suppression.
    }
    \label{tab:compare}
    \vspace{-5pt}
    \begin{tabular}{@{}llcc@{}}
      \toprule
      & Method & Acc. & Text $\Delta$ \\
      \midrule
      & SFT & 89.6\% & +30.5 \\
      \midrule
      Data & Reshuffled options & 89.4\% & +30.2 \\
      \midrule
      \multirow{2}{*}{Loss} & LMH~\citep{clark2019don} & 83.4\% & +13.3 \\
                            & mDPO~\citep{wang2024mdpo} & 87.0\% & +26.1 \\
      \midrule
      Update & Random subspace & 89.6\% & +30.4 \\
      \midrule
      & \textbf{VETO} & 89.2\% & +6.0 \\
      \bottomrule
    \end{tabular}
    
  \end{minipage}%
  \hfill
  \begin{minipage}[t]{0.40\linewidth}
    \centering\footnotesize
    \captionof{table}{Ablations of the constraint with the same columns as in Tab.~\ref{tab:compare}.
    }
    \label{tab:ablation}
    \vspace{-5pt}
    \begin{tabular}{@{}llcc@{}}
      \toprule
      & Setting & Acc. & Text $\Delta$ \\
      \midrule
      & SFT & 89.6\% & +30.5 \\
      \midrule
      \multirow{3}{*}{$\rho$} & 0.90 & 88.6\% & +12.9 \\
                              & \textbf{0.95} & 89.2\% & +6.0 \\
                              & 0.99 & 89.5\% & +1.1 \\
      \midrule
      \multirow{3}{*}{Project} & At init.\ only & 88.7\% & +30.0 \\
                               & After training & 82.4\% & +2.4 \\
                               & \textbf{Every step} & 89.2\% & +6.0 \\
      \bottomrule
    \end{tabular}
    
  \end{minipage}
  \vspace{-10pt}
\end{table}

As the shortcut forms regardless of epoch, data size, or backbone, an intervention can act at the data, the loss, or the update. Tab.~\ref{tab:compare} compares VETO with alternative interventions on Qwen2.5-VL-7B, with each row reporting the accuracy and the text-only gain.

Intervening at the \emph{data} requires knowing which regularity to remove. Reshuffling the answer position every epoch removes the one regularity evident without inspection and leaves the shortcut almost unchanged. With the question fixed, the regularity that remains lies in the text of the options, and a rule that removes it is written for one dataset and does not transfer to the next.

Intervening at the \emph{loss} does not require defining the regularity. Learned-Mixin +H (LMH)~\citep{clark2019don} reduces the gradient where the text alone gives the answer, and multimodal direct preference optimization (mDPO)~\citep{wang2024mdpo} penalizes an answer that stays as likely without the video. LMH suppresses part of the shortcut at a high accuracy cost, and mDPO builds nearly the same shortcut at lower accuracy. Adjusting the loss thus trades accuracy for suppression, whereas VETO keeps the update out of the shortcut and preserves the accuracy.

Intervening at the \emph{update} reaches where the shortcut is written. A random subspace of the same rank leaves the accuracy and text-only accuracy of SFT almost unchanged, so the rank removed is not what matters. What matters is the directions the calibration locates, and removing those directions alone is how VETO keeps the accuracy and suppresses the shortcut with no regularity to define.

Tab.~\ref{tab:ablation} varies the constraint itself. A larger $\rho$ closes more of the input space and suppresses more of the shortcut, and the stronger suppression does not cost accuracy. When the projection is applied is important as well, since projected once at initialization the update drifts back into the subspace and the shortcut forms as under SFT. Projecting once after training suppresses the shortcut at the cost of accuracy, as it discards what was learned along those directions. The shortcut thus accumulates over the steps in between, and only a projection at every step keeps the update out of the subspace throughout. This is what makes VETO preventive rather than corrective.

\section{Related Work}
\label{sec:related}

\textbf{Shortcuts and language priors} have drawn extensive analysis since vision-language models (VLMs) came to rest on LLMs. In the image domain VLMs answer from the question alone~\citep{goyal2017making, agrawal2018don}, and in video many questions are likewise answered from text alone~\citep{yang2025vidlbeval}. The video is read as what typically happens rather than what the clip shows~\citep{buch2022revisiting, lei2023revealing, zhang2024vinoground, liu2024tempcompass}. Fine-tuning then adds what its data teach, as a VLM reproduces what is frequent in its instruction data regardless of the visual input~\citep{li2023evaluating, chen2026language}. Such an added regularity is the shortcut, as distinct from the prior of the base model, and it predicts the label while it may not hold under distribution shift~\citep{geirhos2020shortcut, dancette2021beyond, agrawal2018don}. Remedies remove the bias from the data~\citep{chen2020counterfactual, shvetsova2025unbiasing}, down-weight it in the loss through a question-only model~\citep{cadene2019rubi, clark2019don, niu2021counterfactual} or a text-only penalty against the base model~\citep{chen2026language}, or correct the trained model through preference optimization~\citep{wang2024mdpo} or its decoding through contrastive decoding~\citep{yang2025vidlbeval}. Each first defines the shortcut as a quantity and then removes or penalizes it, but a shortcut outside the definition lies outside the remedy~\citep{dancette2021beyond}. Unlike for images, remedies for video still concentrate on the data and decoding rather than on training itself~\citep{yang2025vidlbeval}. VETO instead constrains the update, keeping it out of the directions in which the shortcut would be learned without defining it, so that the shortcut is suppressed as it forms rather than defined and removed.

\textbf{Constraining the update of fine-tuning} to a subspace separates what a model maintains from what it modifies. Continual learning keeps gradients in the null space of features from earlier tasks~\citep{wang2021training} and the adapter subspaces of successive tasks orthogonal~\citep{wang2023orthogonal} or interference-free~\citep{liang2024inflora}, and model editing projects an edit onto the null space of the keys of knowledge it preserves~\citep{fang2025alphaedit}. The same projection has been applied to representations rather than to updates~\citep{ravfogel2020null, belrose2023leace}, including to remove a text-like subspace from the visual tokens of a VLM~\citep{saini2026language}. In all of these the subspace is defined by what the model already has and specified for it. LoRA~\citep{hu2021lora} gives the constraint its natural object, since its update enters the weights as a few new directions~\citep{shuttleworth2025lora}. VETO then defines the subspace by what fine-tuning would read from the text, so that the same projection prevents rather than preserves. VETO adds only a projection per step and needs no definition of the shortcut, which turns a constraint built to preserve into one that prevents.

\section{Conclusion}
\label{sec:conclusion}

This paper addresses the shortcut SFT builds in a VideoLLM from textual regularities of its training data. The shortcut forms across VideoLLMs, yet it is not what fine-tuning requires for its gain and even costs accuracy under distribution shift. VETO suppresses the shortcut as it forms, preserves the gain of fine-tuning on the training distribution, and gains more than SFT under shift. What VETO calibrates is not a list of regularities but where the text alone is read, so the same projection can make a model use all of its input in other modalities. More broadly, what fine-tuning adds to a VideoLLM is set not only by its data but by the directions its update is allowed to enter, and constraining the update is a means of intervention the data, the loss, and the trained model do not offer.

\section{Limitations and Future Directions}
\label{sec:limitations}

Our analysis of the shortcut rests on multiple-choice video QA, where text-only accuracy can be measured directly. VETO itself is not restricted to this format, but free-form answering has no direct measure of the shortcut yet and is left to future work. The projection of VETO is instantiated with LoRA, and its extension to other adapters on the input side of the update is a further direction. The main experiments fine-tune on SSv2 alone, and the same suppression of the shortcut is observed when fine-tuning on two further datasets, as Appx.~\ref{app:dataset_choice} reports.

\subsection*{AI use statement}

In this work, we used generative AI tools for correcting grammar, formatting tables, and placing figures in the paper. We have not used generative AI tools for preprocessing the training data, which follows the template of each dataset, and all videos and question-answer prompts come from publicly available datasets, with datasets such as Perception Test used under their license for academic research. The formulas in the paper were written by the authors, and generative AI tools were used only to typeset them in LaTeX. Additionally, we used generative AI tools to search for prior work that may overlap with the proposed method, so as to identify if there are any existing works with the same idea. We have reviewed all AI-assisted work. We have checked that every referenced work in the paper is relevant to the point the authors make with it. We take responsibility for the final content of this work, including text, claims or artifacts produced with the aid of generative AI.

\subsection*{Ethics statement}

Video-Evidence Training via Orthogonal projection (VETO) is a training-time technique that constrains how a pretrained VideoLLM is fine-tuned. It operates on existing models and leaves the prior of the base model untouched, so it inherits their capabilities and limitations, including the biases in their pretraining data, and does not correct them. VETO is not a safety filter and does not certify that a model reasons from the video. A model fine-tuned with VETO should therefore not be taken as faithful to its visual input beyond what is measured.

Our experiments rely on publicly available VideoLLMs and on public video datasets used under their licenses, and no new data involving human subjects were collected. This work is intended for academic research. Any application of methods that shape what a model learns during fine-tuning should document the calibration data and constraints used and follow the model and dataset licenses.

\subsection*{Reproducibility statement}

We will make the source code for VETO publicly available, including the calibration, projection, blind filtering, training, and evaluation pipelines together with the data manifests that define every split. The appendix provides the implementation details, including the fine-tuning recipe and hyperparameters, the construction of the calibration set and of the evaluation splits, and the pretrained models and datasets used in our evaluation, and every reported value is a mean over the number of seeds stated there.

\bibliography{iclr2027_conference}

\begin{thebibliography}{45}
\providecommand{\natexlab}[1]{#1}
\providecommand{\url}[1]{\texttt{#1}}
\expandafter\ifx\csname urlstyle\endcsname\relax
  \providecommand{\doi}[1]{doi: #1}\else
  \providecommand{\doi}{doi: \begingroup \urlstyle{rm}\Url}\fi

\bibitem[Agrawal et~al.(2018)Agrawal, Batra, Parikh, and Kembhavi]{agrawal2018don}
Aishwarya Agrawal, Dhruv Batra, Devi Parikh, and Aniruddha Kembhavi.
\newblock Don't just assume; look and answer: Overcoming priors for visual question answering.
\newblock In \emph{2018 IEEE/CVF Conference on Computer Vision and Pattern Recognition}, pp.\  4971--4980. IEEE, 2018.

\bibitem[Bai et~al.(2025{\natexlab{a}})Bai, Cai, Chen, Chen, Chen, Cheng, Deng, Ding, Gao, Ge, Ge, Guo, Huang, Huang, Huang, Hui, Jiang, Li, Li, Li, Li, Lin, Lin, Liu, Liu, Liu, Liu, Liu, Liu, Lu, Luo, Lv, Men, Meng, Ren, Ren, Song, Sun, Tang, Tu, Wan, Wang, Wang, Wang, Wang, Xie, Xu, Xu, Xu, Yang, Yang, Yang, Yang, Yu, Zhang, Zhang, Zhang, Zheng, Zhong, Zhou, Zhou, Zhou, Zhu, and Zhu]{bai2025qwen3vltechnicalreport}
Shuai Bai, Yuxuan Cai, Ruizhe Chen, Keqin Chen, Xionghui Chen, Zesen Cheng, Lianghao Deng, Wei Ding, Chang Gao, Chunjiang Ge, Wenbin Ge, Zhifang Guo, Qidong Huang, Jie Huang, Fei Huang, Binyuan Hui, Shutong Jiang, Zhaohai Li, Mingsheng Li, Mei Li, Kaixin Li, Zicheng Lin, Junyang Lin, Xuejing Liu, Jiawei Liu, Chenglong Liu, Yang Liu, Dayiheng Liu, Shixuan Liu, Dunjie Lu, Ruilin Luo, Chenxu Lv, Rui Men, Lingchen Meng, Xuancheng Ren, Xingzhang Ren, Sibo Song, Yuchong Sun, Jun Tang, Jianhong Tu, Jianqiang Wan, Peng Wang, Pengfei Wang, Qiuyue Wang, Yuxuan Wang, Tianbao Xie, Yiheng Xu, Haiyang Xu, Jin Xu, Zhibo Yang, Mingkun Yang, Jianxin Yang, An~Yang, Bowen Yu, Fei Zhang, Hang Zhang, Xi~Zhang, Bo~Zheng, Humen Zhong, Jingren Zhou, Fan Zhou, Jing Zhou, Yuanzhi Zhu, and Ke~Zhu.
\newblock Qwen3-vl technical report, 2025{\natexlab{a}}.
\newblock URL \url{https://arxiv.org/abs/2511.21631}.

\bibitem[Bai et~al.(2025{\natexlab{b}})Bai, Chen, Liu, Wang, Ge, Song, Dang, Wang, Wang, Tang, Zhong, Zhu, Yang, Li, Wan, Wang, Ding, Fu, Xu, Ye, Zhang, Xie, Cheng, Zhang, Yang, Xu, and Lin]{bai2025qwen25vltechnicalreport}
Shuai Bai, Keqin Chen, Xuejing Liu, Jialin Wang, Wenbin Ge, Sibo Song, Kai Dang, Peng Wang, Shijie Wang, Jun Tang, Humen Zhong, Yuanzhi Zhu, Mingkun Yang, Zhaohai Li, Jianqiang Wan, Pengfei Wang, Wei Ding, Zheren Fu, Yiheng Xu, Jiabo Ye, Xi~Zhang, Tianbao Xie, Zesen Cheng, Hang Zhang, Zhibo Yang, Haiyang Xu, and Junyang Lin.
\newblock Qwen2.5-vl technical report, 2025{\natexlab{b}}.
\newblock URL \url{https://arxiv.org/abs/2502.13923}.

\bibitem[Belrose et~al.(2023)Belrose, Schneider-Joseph, Ravfogel, Cotterell, Raff, and Biderman]{belrose2023leace}
Nora Belrose, David Schneider-Joseph, Shauli Ravfogel, Ryan Cotterell, Edward Raff, and Stella Biderman.
\newblock {LEACE}: Perfect linear concept erasure in closed form.
\newblock In \emph{Thirty-seventh Conference on Neural Information Processing Systems}, 2023.
\newblock URL \url{https://openreview.net/forum?id=awIpKpwTwF}.

\bibitem[Buch et~al.(2022)Buch, Eyzaguirre, Gaidon, Wu, Fei-Fei, and Niebles]{buch2022revisiting}
Shyamal Buch, Crist{\'o}bal Eyzaguirre, Adrien Gaidon, Jiajun Wu, Li~Fei-Fei, and Juan~Carlos Niebles.
\newblock Revisiting the “video” in video-language understanding.
\newblock In \emph{2022 IEEE/CVF Conference on Computer Vision and Pattern Recognition (CVPR)}, pp.\  2907--2917. IEEE, 2022.

\bibitem[Cadene et~al.(2019)Cadene, Dancette, Ben-younes, Cord, and Parikh]{cadene2019rubi}
Remi Cadene, Corentin Dancette, Hedi Ben-younes, Matthieu Cord, and Devi Parikh.
\newblock {RUBi}: Reducing unimodal biases for visual question answering.
\newblock In \emph{Advances in Neural Information Processing Systems}, volume~32, 2019.

\bibitem[Chen et~al.(2020)Chen, Yan, Xiao, Zhang, Pu, and Zhuang]{chen2020counterfactual}
Long Chen, Xin Yan, Jun Xiao, Hanwang Zhang, Shiliang Pu, and Yueting Zhuang.
\newblock Counterfactual samples synthesizing for robust visual question answering.
\newblock In \emph{CVPR}, 2020.

\bibitem[Chen \& Li(2026)Chen and Li]{chen2026language}
Yangneng Chen and Jing Li.
\newblock Language bias in {LVLM}s: From in-depth analysis to simple and effective mitigation.
\newblock In \emph{Forty-third International Conference on Machine Learning}, 2026.
\newblock URL \url{https://openreview.net/forum?id=0tdnyH7uXp}.

\bibitem[Clark et~al.(2019)Clark, Yatskar, and Zettlemoyer]{clark2019don}
Christopher Clark, Mark Yatskar, and Luke Zettlemoyer.
\newblock Don’t take the easy way out: Ensemble based methods for avoiding known dataset biases.
\newblock In \emph{Proceedings of the 2019 conference on empirical methods in natural language processing and the 9th international joint conference on natural language Processing (EMNLP-IJCNLP)}, pp.\  4069--4082, 2019.

\bibitem[Clark et~al.(2026)Clark, Zhang, Ma, Park, Salehi, Tripathi, Lee, Ren, Kim, Yang, Shao, Yang, Huang, Gao, Anderson, Zhang, Jain, Stoica, Han, Farhadi, and Krishna]{clark2026molmo2openweightsdata}
Christopher Clark, Jieyu Zhang, Zixian Ma, Jae~Sung Park, Mohammadreza Salehi, Rohun Tripathi, Sangho Lee, Zhongzheng Ren, Chris~Dongjoo Kim, Yinuo Yang, Vincent Shao, Yue Yang, Weikai Huang, Ziqi Gao, Taira Anderson, Jianrui Zhang, Jitesh Jain, George Stoica, Winson Han, Ali Farhadi, and Ranjay Krishna.
\newblock Molmo2: Open weights and data for vision-language models with video understanding and grounding, 2026.
\newblock URL \url{https://arxiv.org/abs/2601.10611}.

\bibitem[Cores et~al.(2024)Cores, Dorkenwald, Mucientes, Snoek, and Asano]{cores2024lost}
Daniel Cores, Michael Dorkenwald, Manuel Mucientes, Cees~GM Snoek, and Yuki~M Asano.
\newblock Lost in time: A new temporal benchmark for videollms.
\newblock \emph{arXiv preprint arXiv:2410.07752}, 2024.

\bibitem[Dancette et~al.(2021)Dancette, Cadene, Teney, and Cord]{dancette2021beyond}
Corentin Dancette, Remi Cadene, Damien Teney, and Matthieu Cord.
\newblock Beyond question-based biases: Assessing multimodal shortcut learning in visual question answering.
\newblock In \emph{2021 IEEE/CVF International Conference on Computer Vision (ICCV)}, pp.\  1554--1563. IEEE, 2021.

\bibitem[Dao(2024)]{dao2024flashattention}
Tri Dao.
\newblock Flashattention-2: Faster attention with better parallelism and work partitioning.
\newblock In \emph{The Twelfth International Conference on Learning Representations}, 2024.
\newblock URL \url{https://openreview.net/forum?id=mZn2Xyh9Ec}.

\bibitem[Fang et~al.(2025)Fang, Jiang, Wang, Ma, Shi, Wang, He, and Chua]{fang2025alphaedit}
Junfeng Fang, Houcheng Jiang, Kun Wang, Yunshan Ma, Jie Shi, Xiang Wang, Xiangnan He, and Tat-Seng Chua.
\newblock Alphaedit: Null-space constrained knowledge editing for language models.
\newblock In \emph{International Conference on Learning Representations}, volume 2025, pp.\  16366--16396, 2025.

\bibitem[Geirhos et~al.(2020)Geirhos, Jacobsen, Michaelis, Zemel, Brendel, Bethge, and Wichmann]{geirhos2020shortcut}
Robert Geirhos, J{\"o}rn-Henrik Jacobsen, Claudio Michaelis, Richard Zemel, Wieland Brendel, Matthias Bethge, and Felix~A Wichmann.
\newblock Shortcut learning in deep neural networks.
\newblock \emph{Nature Machine Intelligence}, 2\penalty0 (11):\penalty0 665--673, 2020.

\bibitem[Geva et~al.(2023)Geva, Bastings, Filippova, and Globerson]{geva2023dissecting}
Mor Geva, Jasmijn Bastings, Katja Filippova, and Amir Globerson.
\newblock Dissecting recall of factual associations in auto-regressive language models.
\newblock In \emph{Proceedings of the 2023 Conference on Empirical Methods in Natural Language Processing}, pp.\  12216--12235, 2023.

\bibitem[Goyal et~al.(2017{\natexlab{a}})Goyal, Ebrahimi~Kahou, Michalski, Materzynska, Westphal, Kim, Haenel, Fruend, Yianilos, Mueller-Freitag, et~al.]{goyal2017something}
Raghav Goyal, Samira Ebrahimi~Kahou, Vincent Michalski, Joanna Materzynska, Susanne Westphal, Heuna Kim, Valentin Haenel, Ingo Fruend, Peter Yianilos, Moritz Mueller-Freitag, et~al.
\newblock The ``something something'' video database for learning and evaluating visual common sense.
\newblock In \emph{Proceedings of the IEEE international conference on computer vision}, pp.\  5842--5850, 2017{\natexlab{a}}.

\bibitem[Goyal et~al.(2017{\natexlab{b}})Goyal, Khot, Summers-Stay, Batra, and Parikh]{goyal2017making}
Yash Goyal, Tejas Khot, Douglas Summers-Stay, Dhruv Batra, and Devi Parikh.
\newblock Making the v in vqa matter: Elevating the role of image understanding in visual question answering.
\newblock In \emph{Proceedings of the IEEE conference on computer vision and pattern recognition}, pp.\  6904--6913, 2017{\natexlab{b}}.

\bibitem[Hu et~al.(2021)Hu, Shen, Wallis, Allen-Zhu, Li, Wang, Wang, and Chen]{hu2021lora}
Edward~J Hu, Yelong Shen, Phillip Wallis, Zeyuan Allen-Zhu, Yuanzhi Li, Shean Wang, Lu~Wang, and Weizhu Chen.
\newblock Lora: Low-rank adaptation of large language models.
\newblock \emph{arXiv preprint arXiv:2106.09685}, 2021.

\bibitem[Lei et~al.(2023)Lei, Berg, and Bansal]{lei2023revealing}
Jie Lei, Tamara Berg, and Mohit Bansal.
\newblock Revealing single frame bias for video-and-language learning.
\newblock In \emph{Proceedings of the 61st Annual Meeting of the Association for Computational Linguistics (Volume 1: Long Papers)}, pp.\  487--507, 2023.

\bibitem[Leng et~al.(2024)Leng, Zhang, Chen, Li, Lu, Miao, and Bing]{leng2024mitigating}
Sicong Leng, Hang Zhang, Guanzheng Chen, Xin Li, Shijian Lu, Chunyan Miao, and Lidong Bing.
\newblock Mitigating object hallucinations in large vision-language models through visual contrastive decoding.
\newblock In \emph{2024 IEEE/CVF Conference on Computer Vision and Pattern Recognition (CVPR)}, pp.\  13872--13882. IEEE, 2024.

\bibitem[Li et~al.(2025)Li, Zhang, Guo, Zhang, Li, Zhang, Zhang, Zhang, Li, Liu, and Li]{li2025llavaonevision}
Bo~Li, Yuanhan Zhang, Dong Guo, Renrui Zhang, Feng Li, Hao Zhang, Kaichen Zhang, Peiyuan Zhang, Yanwei Li, Ziwei Liu, and Chunyuan Li.
\newblock {LL}a{VA}-onevision: Easy visual task transfer.
\newblock \emph{Transactions on Machine Learning Research}, 2025.
\newblock ISSN 2835-8856.
\newblock URL \url{https://openreview.net/forum?id=zKv8qULV6n}.

\bibitem[Li et~al.(2024)Li, Wang, He, Li, Wang, Liu, Wang, Xu, Chen, Luo, et~al.]{li2024mvbench}
Kunchang Li, Yali Wang, Yinan He, Yizhuo Li, Yi~Wang, Yi~Liu, Zun Wang, Jilan Xu, Guo Chen, Ping Luo, et~al.
\newblock Mvbench: A comprehensive multi-modal video understanding benchmark.
\newblock In \emph{Proceedings of the IEEE/CVF conference on computer vision and pattern recognition}, pp.\  22195--22206, 2024.

\bibitem[Li et~al.(2023)Li, Du, Zhou, Wang, Zhao, and Wen]{li2023evaluating}
Yifan Li, Yifan Du, Kun Zhou, Jinpeng Wang, Xin Zhao, and Ji-Rong Wen.
\newblock Evaluating object hallucination in large vision-language models.
\newblock In \emph{Proceedings of the 2023 conference on empirical methods in natural language processing}, pp.\  292--305, 2023.

\bibitem[Liang \& Li(2024)Liang and Li]{liang2024inflora}
Yan-Shuo Liang and Wu-Jun Li.
\newblock Inflora: Interference-free low-rank adaptation for continual learning.
\newblock In \emph{2024 IEEE/CVF Conference on Computer Vision and Pattern Recognition (CVPR)}, pp.\  23638--23647. IEEE, 2024.

\bibitem[Liu et~al.(2024)Liu, Li, Liu, Wang, Ren, Li, Chen, Sun, and Hou]{liu2024tempcompass}
Yuanxin Liu, Shicheng Li, Yi~Liu, Yuxiang Wang, Shuhuai Ren, Lei Li, Sishuo Chen, Xu~Sun, and Lu~Hou.
\newblock Tempcompass: Do video llms really understand videos?
\newblock In \emph{Findings of the Association for Computational Linguistics: ACL 2024}, pp.\  8731--8772, 2024.

\bibitem[Loshchilov \& Hutter(2017)Loshchilov and Hutter]{loshchilov2017decoupled}
Ilya Loshchilov and Frank Hutter.
\newblock Decoupled weight decay regularization.
\newblock \emph{arXiv preprint arXiv:1711.05101}, 2017.

\bibitem[Niu et~al.(2021)Niu, Tang, Zhang, Lu, Hua, and Wen]{niu2021counterfactual}
Yulei Niu, Kaihua Tang, Hanwang Zhang, Zhiwu Lu, Xian-Sheng Hua, and Ji-Rong Wen.
\newblock Counterfactual vqa: A cause-effect look at language bias.
\newblock In \emph{2021 IEEE/CVF Conference on Computer Vision and Pattern Recognition (CVPR)}, pp.\  12695--12705. IEEE, 2021.

\bibitem[nostalgebraist(2020)]{nostalgebraist2020logitlens}
nostalgebraist.
\newblock interpreting {GPT}: the logit lens, 2020.
\newblock URL \url{https://www.lesswrong.com/posts/AcKRB8wDpdaN6v6ru/interpreting-gpt-the-logit-lens}.
\newblock LessWrong blog post.

\bibitem[Patraucean et~al.(2023)Patraucean, Smaira, Gupta, Continente, Markeeva, Banarse, Koppula, Heyward, Malinowski, Yang, Doersch, Matejovicova, Sulsky, Miech, Fr{\'e}chette, Klimczak, Koster, Zhang, Winkler, Aytar, Osindero, Damen, Zisserman, and Carreira]{patraucean2023perception}
Viorica Patraucean, Lucas Smaira, Ankush Gupta, Adria~Recasens Continente, Larisa Markeeva, Dylan~Sunil Banarse, Skanda Koppula, Joseph Heyward, Mateusz Malinowski, Yi~Yang, Carl Doersch, Tatiana Matejovicova, Yury Sulsky, Antoine Miech, Alexandre Fr{\'e}chette, Hanna Klimczak, Raphael Koster, Junlin Zhang, Stephanie Winkler, Yusuf Aytar, Simon Osindero, Dima Damen, Andrew Zisserman, and Joao Carreira.
\newblock Perception test: A diagnostic benchmark for multimodal video models.
\newblock In \emph{Thirty-seventh Conference on Neural Information Processing Systems Datasets and Benchmarks Track}, 2023.
\newblock URL \url{https://openreview.net/forum?id=HYEGXFnPoq}.

\bibitem[Ravfogel et~al.(2020)Ravfogel, Elazar, Gonen, Twiton, and Goldberg]{ravfogel2020null}
Shauli Ravfogel, Yanai Elazar, Hila Gonen, Michael Twiton, and Yoav Goldberg.
\newblock Null it out: Guarding protected attributes by iterative nullspace projection.
\newblock In \emph{Proceedings of the 58th annual meeting of the association for computational linguistics}, pp.\  7237--7256, 2020.

\bibitem[Saini et~al.(2026)Saini, Jha, Tang, and Liu]{saini2026language}
Harshvardhan Saini, Samyak Jha, Yiming Tang, and Dianbo Liu.
\newblock When language overwrites vision: Over-alignment and geometric debiasing in vision-language models.
\newblock \emph{arXiv preprint arXiv:2605.08245}, 2026.

\bibitem[Shuttleworth et~al.(2025)Shuttleworth, Andreas, Torralba, and Sharma]{shuttleworth2025lora}
Reece~S Shuttleworth, Jacob Andreas, Antonio Torralba, and Pratyusha Sharma.
\newblock Lo{RA} vs full fine-tuning: An illusion of equivalence.
\newblock In \emph{The Thirty-ninth Annual Conference on Neural Information Processing Systems}, 2025.
\newblock URL \url{https://openreview.net/forum?id=xp7B8rkh7L}.

\bibitem[Shvetsova et~al.(2025)Shvetsova, Nagrani, Schiele, Kuehne, and Rupprecht]{shvetsova2025unbiasing}
Nina Shvetsova, Arsha Nagrani, Bernt Schiele, Hilde Kuehne, and Christian Rupprecht.
\newblock Unbiasing through textual descriptions: Mitigating representation bias in video benchmarks.
\newblock In \emph{2025 IEEE/CVF Conference on Computer Vision and Pattern Recognition (CVPR)}, pp.\  29050--29059. IEEE, 2025.

\bibitem[Team et~al.(2026)Team, Abd, Aggarwal, Algayres, Andreev, Bachem, Ballantyne, Brick, Cărbune, Casbon, Chaturvedi, Chawla, Cotruta, Coucke, Culliton, Dadashi, Dixon, Elhawaty, Evci, Farabet, Ferret, Galgani, Girgin, Grill, Grootendorst, Guo, Hardin, He, Hernandez, Homburger, Hussenot, Ji, Joulin, Kamath, Kassraie, Lacombe, Lahoti, Liu, Martins, Martins, Matejovicova, Merhej, Momchev, Mondal, Mullins, Panyam, Pathak, Perrin, Pinto, Pot, Pouget, Ramé, Ramos, Reid, Rim, Rivière, Roth, Rouillard, Sanseviero, Sessa, Settle, Sinopalnikov, Smoot, Stanczyk, Steiner, Stewart, Tolstikhin, Tschannen, Tsitsulin, Vieillard, Wu, Xu, Yang, Yvinec, Zhang, Zhang, Zou, Aagnes, Abdelhamed, Adamek, Agrawal, Agrawal, Alabdulmohsin, Alayrac, Alon, Amarnath, Anand, Anastasiou, Ariafar, Aubet, Axiotis, Barbero, Barral, Bendebury, Bergmann, Bileschi, Black, Blondel, Borgeaud, Bražinskas, Burnell, Busa-Fekete, Cai, Calandriello, Cameron, Caucheteux, Chaabouni, Chadha, Chan, Chen, Chen, Chen, Chen, Cheng, hsiang Chien,
  Chinaev, Chou, Chu, Coleman, Consul, Conway-Rahman, Crowell, Cutler, Dani, Daruki, Das, Deutsch, Dikkala, Ding, Ding, Dodhia, Donhauser, Doshi, Dragan, Druinsky, Dua, Egyed, Eisenbud, Eppens, Fan, Fatemi, Fathullah, Feinberg, Ferev, Flennerhag, Fujimoto, Oliveira, Galatzer-Levy, Gante, Geisler, Ghosal, Girgis, von Glehn, Go, Gokhale, Grills, Gu, Gupta, Gupta, Guruganesh, Hadsell, Harkous, Harlalka, Hassabis, Hauth, Heyward, Hosseini, Hsia, Hsu, Huang, Huang, Hui, Hutter, I, Iliopoulos, Jain, Jawahar, Ji, Jin, Johnson, Joshi, Kandoor, Kang, Kavukcuoglu, Kazemi, Kenealy, Khalifa, Kirk, Korotkov, Kothawade, Kovalev, Kovelamudi, Kraft, Kumar, Kumar, Kuppam, Lannin, Lee, Lee, Lepikhin, Levkovitch, Li, Li, Liévin, Lin, Lin, Liu, Liu, Liu, Liu, Lobov, Lunayach, Ma, Madan, Maksai, Malmi, Matuszak, McDuff, Menghani, Mikuła, Mirylenka, Misiunas, Misra, Mitran, Mohamed, Mukha, Noland, O'Donnell, O'Donoghue, Olszewska, Orlando, Pan, Panigrahy, Parekh, Perez-Nieves, Park, Paskie, Peng, Petrini, Petrov, Pfeiffer, Piot,
  Plomecka, Poder, Ponce, Pramanik, Racz, Rajan, Ramanovich, Rao, Ritter, Rodrigues, Rosen, Rybiński, Sachdeva, Sander, Sathyanarayana, Savla, Schmidgall, Schuster, Scrivener, Seguin, Sellergren, Severyn, Shafran, Shah, Shahriari, Shangguan, Shenoy, Shenoy, Shivanna, Sho, Spangher, Stokowiec, Strother, Su, Sun, Sundararajan, Tacchetti, Taege, Tafti, Tarbouriech, Tekur, Thakoor, Thapa, Traverse, Treven, Tu, Tung, Çağlar Ünlü, Veličković, Venkat, Venkatesh, Venkiteswaran, Visin, Vitvitskyi, Vodrahalli, Wang, Wang, Warkentin, Wassenberg, Wieting, Wu, Xiao, Xu, Xu, Xue, Yadav, Yan, Yang, Yang, Yang, Ying, Yoo, Zadimoghaddam, Zafar, Zhang, Zhang, Zhang, Zhang, Zhao, Zhou, and Zou]{gemmateam2026gemma4technicalreport}
Gemma Team, Sherif~El Abd, Vaibhav Aggarwal, Robin Algayres, Alek Andreev, Olivier Bachem, Ian Ballantyne, Cormac Brick, Victor Cărbune, Michelle Casbon, Mayank Chaturvedi, Aditya Chawla, Victor Cotruta, Alice Coucke, Phil Culliton, Robert Dadashi, Lucas Dixon, Mohamed Elhawaty, Utku Evci, Clément Farabet, Johan Ferret, Filippo Galgani, Sertan Girgin, Jean-Bastien Grill, Maarten Grootendorst, Jiaxian Guo, Cassidy Hardin, Yanzhang He, Steven~M. Hernandez, Omri Homburger, Léonard Hussenot, Juyeong Ji, Armand Joulin, Aishwarya Kamath, Parnian Kassraie, Olivier Lacombe, Preethi Lahoti, Gaël Liu, Gus Martins, Luciano Martins, Tatiana Matejovicova, Ramona Merhej, Nikola Momchev, Sneha Mondal, Ryan Mullins, Sindhu~Raghuram Panyam, Shreya Pathak, Sarah Perrin, André~Susano Pinto, Etienne Pot, Angéline Pouget, Alexandre Ramé, Sabela Ramos, Douglas Reid, David Rim, Morgane Rivière, Karsten Roth, Louis Rouillard, Omar Sanseviero, Pier~Giuseppe Sessa, Shane Settle, Danila Sinopalnikov, Sara Smoot, Piotr Stanczyk,
  Andreas Steiner, Lawrence Stewart, Ilya Tolstikhin, Michael Tschannen, Anton Tsitsulin, Nino Vieillard, Renjie Wu, Pingmei Xu, Haichuan Yang, Edouard Yvinec, Biao Zhang, Li~Zhang, Joe Zou, Nicolas Aagnes, Abdelrahman Abdelhamed, Jakub Adamek, Shivani Agrawal, Shubham Agrawal, Ibrahim Alabdulmohsin, Jean~Baptiste Alayrac, Uri Alon, Chandramouli Amarnath, Ankesh Anand, Chrysovalantis Anastasiou, Setareh Ariafar, François-Xavier Aubet, Kyriakos Axiotis, Federico Barbero, Joelle Barral, Alexei Bendebury, Urs Bergmann, Stanley Bileschi, Kat Black, Mathieu Blondel, Sebastian Borgeaud, Arthur Bražinskas, Ryan Burnell, Robert Busa-Fekete, Mu~Cai, Daniele Calandriello, Glenn Cameron, Charlotte Caucheteux, Rahma Chaabouni, Garima Chadha, Jetha Chan, Blake~Jianhang Chen, Jesse Chen, Lin Chen, Xu~Chen, Derek Cheng, Tzu hsiang Chien, Nikolai Chinaev, Yi~Chou, Zhaohui Chu, Benjamin Coleman, Pooja Consul, Sam Conway-Rahman, Scott Crowell, Dylan Cutler, Vivek Dani, Samira Daruki, Anil Das, Daniel Deutsch, Nishanth
  Dikkala, Li~Ding, Qiuhan Ding, Shenil Dodhia, Konstantin Donhauser, Tulsee Doshi, Anca Dragan, Alex Druinsky, Sahil Dua, Zoltan Egyed, Danielle Eisenbud, Daniel Eppens, Cindy Fan, Bahare Fatemi, Yassir Fathullah, Vlad Feinberg, Milen Ferev, Sebastian Flennerhag, Takumi Fujimoto, João~Gabriel Oliveira, Isaac Galatzer-Levy, João Gante, Simon Geisler, Soham Ghosal, Antonious~M. Girgis, Tamara von Glehn, Alec Go, Alhaad Gokhale, Alex Grills, Yiming Gu, Mayank Gupta, Pramod Gupta, Guru Guruganesh, Raia Hadsell, Hamza Harkous, Jitendra Harlalka, Demis Hassabis, Anja Hauth, Joe Heyward, Arian Hosseini, Chih-Yang Hsia, I-Hung Hsu, Xiaopeng Huang, Yangsibo Huang, Kevin Hui, Adrian Hutter, Te~I, Fotis Iliopoulos, Advait Jain, Ganesh Jawahar, Ziwei Ji, Qilin Jin, Melvin Johnson, Kandarp Joshi, Arun Kandoor, Wang-Cheng Kang, Koray Kavukcuoglu, Mehran Kazemi, Kathleen Kenealy, Amr Khalifa, Phoebe Kirk, Ivan Korotkov, Suraj Kothawade, Vitaly Kovalev, Neel Kovelamudi, Adam Kraft, Ravin Kumar, Vivek Kumar, Harish Kuppam,
  Justin Lannin, Chen-Yu Lee, Seungji Lee, Dmitry Lepikhin, Alon Levkovitch, Dongdong Li, Qiujia Li, Valentin Liévin, Ethan Lin, Ziqian Lin, Casper Liu, Tianlin Liu, Tianqi Liu, Xin Liu, Ivan Lobov, Mayank Lunayach, Min Ma, Gagan Madan, Andrii Maksai, Eric Malmi, Michal Matuszak, Daniel McDuff, Gaurav Menghani, Maciej Mikuła, Daniil Mirylenka, Karolis Misiunas, Vedant Misra, Andreea Mitran, Kareem Mohamed, Maksim Mukha, Eric Noland, James O'Donnell, Brendan O'Donoghue, Kate Olszewska, Bernett Orlando, Wanqiong Pan, Rina Panigrahy, Unnati Parekh, Nicolas Perez-Nieves, Chunjong Park, Eric Paskie, Liqian Peng, Bryce Petrini, Slav Petrov, Jonas Pfeiffer, Bilal Piot, Martyna Plomecka, Siim Poder, Octavio Ponce, Arijit Pramanik, David Racz, Anish Rajan, Michelle Ramanovich, Anand Rao, Marvin Ritter, Vitor Rodrigues, Evan Rosen, Mikołaj Rybiński, Noveen Sachdeva, Michaël~E. Sander, Rohit Sathyanarayana, Sagar Savla, Samuel Schmidgall, Tal Schuster, George Scrivener, Benoit Seguin, Andrew Sellergren, Aliaksei
  Severyn, Izhak Shafran, Dhruv Shah, Bobak Shahriari, Yuan Shangguan, Ashish Shenoy, Pradeep Shenoy, Rakesh Shivanna, Pauline Sho, Lucas Spangher, Wojciech Stokowiec, Tim Strother, Yao Su, Yinghao Sun, Mukund Sundararajan, Andrea Tacchetti, Mor~Hazan Taege, Pouya Tafti, Jean Tarbouriech, Chetan Tekur, Shantanu Thakoor, Rahul Thapa, Madeleine Traverse, Lenart Treven, Tao Tu, Chien~Te Tung, Çağlar Ünlü, Petar Veličković, Malini~Pooni Venkat, Sagar~Gubbi Venkatesh, Vidya Venkiteswaran, Francesco Visin, Alex Vitvitskyi, Kiran Vodrahalli, Weiyi Wang, Xin Wang, Tris Warkentin, Jan Wassenberg, John Wieting, Cindy Wu, Lechao Xiao, Hao Xu, Yuhui Xu, Fuzhao Xue, Arun Yadav, Jun Yan, Antoine Yang, Lin Yang, Ming-Hsuan Yang, Ziyu Ying, Jae~Hyeon Yoo, Morteza Zadimoghaddam, Sajjad Zafar, Fred Zhang, Jiageng Zhang, Jianyi Zhang, Xiaofan Zhang, Chao Zhao, David Zhou, and Chen Zou.
\newblock Gemma 4 technical report, 2026.
\newblock URL \url{https://arxiv.org/abs/2607.02770}.

\bibitem[Wang et~al.(2024)Wang, Zhou, Huang, Xu, Zhang, Poon, and Chen]{wang2024mdpo}
Fei Wang, Wenxuan Zhou, James~Y Huang, Nan Xu, Sheng Zhang, Hoifung Poon, and Muhao Chen.
\newblock mdpo: Conditional preference optimization for multimodal large language models.
\newblock In \emph{Proceedings of the 2024 Conference on Empirical Methods in Natural Language Processing}, pp.\  8078--8088, 2024.

\bibitem[Wang et~al.(2025{\natexlab{a}})Wang, Yu, Yuan, Mao, and Zhou]{wang2025videorft}
Qi~Wang, Yanrui Yu, Ye~Yuan, Rui Mao, and Tianfei Zhou.
\newblock Video{RFT}: Incentivizing video reasoning capability in {MLLM}s via reinforced fine-tuning.
\newblock In \emph{The Thirty-ninth Annual Conference on Neural Information Processing Systems}, 2025{\natexlab{a}}.
\newblock URL \url{https://openreview.net/forum?id=3pORFyKzh1}.

\bibitem[Wang et~al.(2021)Wang, Li, Sun, and Xu]{wang2021training}
Shipeng Wang, Xiaorong Li, Jian Sun, and Zongben Xu.
\newblock Training networks in null space of feature covariance for continual learning.
\newblock In \emph{2021 IEEE/CVF Conference on Computer Vision and Pattern Recognition (CVPR)}, pp.\  184--193. IEEE, 2021.

\bibitem[Wang et~al.(2023)Wang, Chen, Ge, Xia, Bao, Zheng, Zhang, Gui, and Huang]{wang2023orthogonal}
Xiao Wang, Tianze Chen, Qiming Ge, Han Xia, Rong Bao, Rui Zheng, Qi~Zhang, Tao Gui, and Xuan-Jing Huang.
\newblock Orthogonal subspace learning for language model continual learning.
\newblock In \emph{Findings of the Association for Computational Linguistics: EMNLP 2023}, pp.\  10658--10671, 2023.

\bibitem[Wang et~al.(2025{\natexlab{b}})Wang, Yoon, Yu, Islam, Bertasius, and Bansal]{wang2025video}
Ziyang Wang, Jaehong Yoon, Shoubin Yu, Md~Mohaiminul Islam, Gedas Bertasius, and Mohit Bansal.
\newblock Video-rts: Rethinking reinforcement learning and test-time scaling for efficient and enhanced video reasoning.
\newblock In \emph{Proceedings of the 2025 Conference on Empirical Methods in Natural Language Processing}, pp.\  28126--28140, 2025{\natexlab{b}}.

\bibitem[Wu et~al.(2021)Wu, Yu, Chen, Tenenbaum, and Gan]{wu2021star}
Bo~Wu, Shoubin Yu, Zhenfang Chen, Joshua~B. Tenenbaum, and Chuang Gan.
\newblock {STAR}: A benchmark for situated reasoning in real-world videos.
\newblock In \emph{Thirty-fifth Conference on Neural Information Processing Systems Datasets and Benchmarks Track (Round 2)}, 2021.
\newblock URL \url{https://openreview.net/forum?id=EfgNF5-ZAjM}.

\bibitem[Xiao et~al.(2021)Xiao, Shang, Yao, and Chua]{xiao2021next}
Junbin Xiao, Xindi Shang, Angela Yao, and Tat-Seng Chua.
\newblock Next-qa: Next phase of question-answering to explaining temporal actions.
\newblock In \emph{2021 IEEE/CVF Conference on Computer Vision and Pattern Recognition (CVPR)}, pp.\  9772--9781. IEEE, 2021.

\bibitem[Xue et~al.(2025)Xue, Luo, and Grauman]{xue2025seeing}
Zihui Xue, Mi~Luo, and Kristen Grauman.
\newblock Seeing the arrow of time in large multimodal models.
\newblock In \emph{The Thirty-ninth Annual Conference on Neural Information Processing Systems}, 2025.
\newblock URL \url{https://openreview.net/forum?id=OYciB30Z4n}.

\bibitem[Yang et~al.(2025)Yang, Guo, Lu, and Wang]{yang2025vidlbeval}
Yiming Yang, Yangyang Guo, Hui Lu, and Yan Wang.
\newblock Vidlbeval: Benchmarking and mitigating language bias in video-involved lvlms.
\newblock \emph{arXiv preprint arXiv:2502.16602}, 2025.

\bibitem[Zhang et~al.(2024)Zhang, Cai, and Lee]{zhang2024vinoground}
Jianrui Zhang, Mu~Cai, and Yong~Jae Lee.
\newblock Vinoground: Scrutinizing lmms over dense temporal reasoning with short videos.
\newblock \emph{arXiv preprint arXiv:2410.02763}, 2024.

\end{thebibliography}
\bibliographystyle{iclr2027_conference}

\appendix
\setcounter{table}{0}\renewcommand{\thetable}{\Alph{table}}
\setcounter{figure}{0}\renewcommand{\thefigure}{\Alph{figure}}
\newpage
\section*{Appendix}

\section{SSv2 Family Construction}
\label{app:families}

Of the 174 SSv2 action templates~\citep{goyal2017something}, 47 come in hard-negative families of two to four templates. The templates of a family share the same object placeholders and, once rendered for a given clip, differ only in temporal content. We use these 47 in 16 families. Families fall into three tiers by the kind of temporal evidence they require. Direction families (horizontal, vertical, relative, two camera-motion families, object-relative) require the direction of motion. State-transition families (containment, cover, open-close) require the order of two states. Outcome families (lift, fall, spin, slant, poke, push, plug) require the result of an action that is visible only at the end of the clip.

Each clip yields one multiple-choice question. The correct option is the template of the clip, and the distractors are the sibling templates re-rendered with the placeholders of that clip. For the eight two-template families we pad to four options with designated near-family distractors, which serve as options and never as answer sources, and the three-template family keeps its three options. Tab.~\ref{tab:families} lists every family with its templates. The retained templates cover roughly 38\% of SSv2 clips.

\begin{table}[h]
  \centering
  \caption{The SSv2 families with their number of templates, items in the blind-filtered training pool and evaluation split, and templates in order of frequency. [X] marks the object placeholders of SSv2.
  }
  \label{tab:families}
  \scriptsize
  \setlength{\tabcolsep}{3pt}
  \begin{tabular}{@{}l r r r p{\dimexpr\linewidth-4.9cm\relax}@{}}
    \toprule
    Family & Templ. & Train & Eval & Templates \\
    \midrule
    Horizontal & 4 & 8,562 & 756 & Pushing [X] from left to right / Pushing [X] from right to left / Pulling [X] from right to left / Pulling [X] from left to right \\
    Vertical & 4 & 7,786 & 789 & Moving [X] up / Moving [X] down / Lifting [X] up completely without letting it drop down / Lifting [X] up completely, then letting it drop down \\
    Relative motion & 4 & 3,507 & 483 & Moving [X] and [X] away from each other / Moving [X] and [X] closer to each other / Moving [X] and [X] so they pass each other / Moving [X] and [X] so they collide with each other \\
    Camera, horizontal & 4 & 3,495 & 419 & Moving away from [X] with your camera / Turning the camera left while filming [X] / Turning the camera right while filming [X] / Approaching [X] with your camera \\
    Camera, vertical & 4 & 3,136 & 266 & Moving [X] away from the camera / Turning the camera downwards while filming [X] / Turning the camera upwards while filming [X] / Moving [X] toward the camera \\
    Object-relative & 2 & 1,753 & 383 & Moving [X] away from [X] / Moving [X] closer to [X] \\
    Containment & 4 & 5,438 & 874 & Putting [X] into [X] / Taking [X] out of [X] / Stuffing [X] into [X] / Pulling [X] out of [X] \\
    Cover & 2 & 4,761 & 656 & Covering [X] with [X] / Uncovering [X] \\
    Open--close & 2 & 2,053 & 480 & Closing [X] / Opening [X] \\
    Lift & 2 & 2,829 & 170 & Lifting up one end of [X] without letting it drop down / Lifting up one end of [X], then letting it drop down \\
    Fall & 2 & 2,446 & 680 & [X] falling like a rock / [X] falling like a feather or paper \\
    Spin & 2 & 1,765 & 452 & Spinning [X] that quickly stops spinning / Spinning [X] so it continues spinning \\
    Slant & 2 & 638 & 88 & Putting [X] that can't roll onto a slanted surface, so it slides down / Putting [X] that can't roll onto a slanted surface, so it stays where it is \\
    Poke & 4 & 3,511 & 378 & Poking [X] so lightly that it doesn't or almost doesn't move / Poking [X] so it slightly moves / Poking [X] so that it falls over / Poking [X] so that it spins around \\
    Push & 3 & 3,879 & 618 & Pushing [X] so that it slightly moves / Pushing [X] so that it falls off the table / Pushing [X] so that it almost falls off but doesn't \\
    Plug & 2 & 1,363 & 524 & Plugging [X] into [X] / Plugging [X] into [X] but pulling it right out as you remove your hand \\
    \midrule
    Total & 47 & 56,922 & 8,016 & \\
    \bottomrule
  \end{tabular}
\end{table}

\section{Blind Filtering Details}
\label{app:filtering}

We run the base model on each split under the text-only input three times with the option order shuffled, and discard items it answers correctly in all three. The rule keeps an item whenever the base model fails at least once in three, and the base model answers 19.3\% of the kept items from the text, below chance. The filter removes 10.4\% of the training pool, 63,526 to 56,922, and 10.7\% of the evaluation pool, 8,977 to 8,016. The same filter applied to the benchmarks of Sec.~\ref{sec:expsetup} keeps 5,637 of the 7,392 multiple-choice items of the released training split of Perception Test~\citep{patraucean2023perception}, 665 of the 1,000 of Vinoground~\citep{zhang2024vinoground}, 1,350 of the 1,757 valid items of AoTBench~\citep{xue2025seeing}, 2,727 of the 3,586 of MVBench~\citep{li2024mvbench}, 1,767 of the 2,205 of TVBench~\citep{cores2024lost}, and 1,097 of the 1,580 of TempCompass~\citep{liu2024tempcompass}. Every result under the shift is reported on these subsets.

Throughout the paper, chance refers to uniform guessing over the options, $0.257$ on the SSv2 evaluation split since one family has three options. The answer letters of the split are nearly balanced, so guessing by their empirical frequencies gives $0.257$ and always choosing the most frequent letter gives $0.266$. On the blind-filtered SSv2 evaluation split the base model reaches $0.193$ under the text-only input, below all three. On every filtered set the text-only accuracy of the base model lies below uniform chance by construction of the filter, so Text $\Delta$ is measured against the base model rather than against chance. Under VETO the text-only accuracy on the SSv2 evaluation split returns to chance, $0.253$ against $0.257$, whereas under SFT it rises to twice chance.

The blind filter uses the base model only, so it takes the language prior of the base model out of the measurement and nothing else. Any text-only accuracy above that of the base model after fine-tuning is therefore added by fine-tuning, and this addition is the shortcut of Sec.~\ref{sec:setup}.

\section{Calibration Set Size}
\label{app:calib}

The shortcut subspace of Sec.~\ref{sec:subspace} is estimated from a calibration set, and this section asks how many items the estimate needs, how much it depends on which items are drawn, and how the energy fraction $\rho$ sets its rank. Coverage plateaus at 400 stratified items, and $\rho = 0.95$ is the last value at which the subspace stays small, and the paper uses these two. We measure \emph{coverage}, the fraction of text-only activation energy of $1{,}000$ held-out items that falls inside $U_1^l$, that is, $\mathrm{tr}\big((U_1^l)^\top C_{\text{held}}^l U_1^l\big) / \mathrm{tr}(C_{\text{held}}^l)$ averaged over all adapted modules. A subspace that generalizes perfectly would reach the $\rho = 0.95$ imposed on the calibration items, and a random subspace of the same rank would reach $r^l / d_{\text{in}} \approx 0.07$. Between two subspaces estimated from different draws we measure \emph{overlap}, $\|(U_a^l)^\top U_b^l\|_F^2 / \min(r_a^l, r_b^l)$, which equals $1$ when one subspace contains the other and again $r^l / d_{\text{in}}$ for two random subspaces.

\begin{table}[t]
  \centering
  \caption{Calibration size sweep, Qwen2.5-VL-7B, three seeds per row, mean $\pm$ sd over seeds. Random rows draw $n$ items from the training pool as nested prefixes of a shuffled order, and stratified rows draw $n/16$ items per family, which is how the calibration set of the paper is drawn. Coverage is measured on $1{,}000$ held-out items, overlap is averaged over the three seed pairs, and the random baseline for both is $r^l/d_{\text{in}} \approx 0.07$.}
  \label{tab:calib_size}
  \small
  \begin{tabular}{@{}lrccc@{}}
    \toprule
    Draw & $n$ & Coverage & Overlap between seeds & Rank $\bar{r}$ \\
    \midrule
    random & 10    & $0.8694 \pm 0.0051$ & 0.63 & 103 \\
    random & 50    & $0.9205 \pm 0.0062$ & 0.78 & 171 \\
    random & 100   & $0.9299 \pm 0.0071$ & 0.83 & 195 \\
    random & 200   & $0.9387 \pm 0.0024$ & 0.88 & 217 \\
    random & 1,000 & $0.9471 \pm 0.0002$ & 0.95 & 246 \\
    random & 2,000 & $0.9484 \pm 0.0003$ & 0.97 & 251 \\
    random & 4,000 & $0.9487 \pm 0.0001$ & 0.98 & 252 \\
    \midrule
    stratified & 200   & $0.9444 \pm 0.0003$ & 0.90 & 241 \\
    \textbf{stratified (used)} & \textbf{400} & $\mathbf{0.9485 \pm 0.0005}$ & 0.93 & 253 \\
    stratified & 500   & $0.9485 \pm 0.0005$ & 0.94 & 251 \\
    stratified & 600   & $0.9496 \pm 0.0001$ & 0.94 & 257 \\
    stratified & 800   & $0.9499 \pm 0.0004$ & 0.95 & 256 \\
    stratified & 1,000 & $0.9504 \pm 0.0005$ & 0.96 & 258 \\
    \bottomrule
  \end{tabular}
\end{table}

Tab.~\ref{tab:calib_size} sweeps the calibration size from 10 to $4{,}000$ randomly drawn items for Qwen2.5-VL-7B with three seeds, where each seed shuffles the training pool once and the sets are nested prefixes of that order, so that the comparison across sizes is not confounded by which items are drawn. Coverage rises quickly and then saturates, with 50 items already at $0.9205$, $1{,}000$ items at $0.947$, and the next $3{,}000$ items adding $0.002$, and the standard deviation over seeds stays below $0.01$ at every size. The estimate is thus a property of the backbone rather than of the items.

The lower half of Tab.~\ref{tab:calib_size} repeats the sweep with draws stratified over the 16 families, which is how the calibration set of the paper is drawn, and stratification lifts the whole curve, as 200 stratified items reach $0.9444$ where 200 random items reach $0.9387$. Along the stratified curve the gain from 200 to 400 items is $0.004$, whereas the gain from 400 to $1{,}000$ items is $0.002$ in total, and the three seeds at any size differ by at most $0.0005$, so the curve plateaus at 400 items. We take the calibration size at the onset of this plateau, 400 items, which sit within $0.002$ of the $\rho$ imposed on the calibration items, and the rank retained there is within five dimensions of the rank at $1{,}000$ items. Overlap between seeds keeps rising past the plateau, from $0.93$ at 400 items to $0.96$ at $1{,}000$, and since coverage does not rise with it, the items beyond the plateau settle which of the near-degenerate tail directions of the calibration spectrum enter $U_1^l$ rather than how much shortcut energy it captures.

\begin{wraptable}[12]{r}{0.38\linewidth}
  \vspace{-\baselineskip}
  \centering
  \caption{Retained rank and held-out coverage against $\rho$.}
  \label{tab:rho}
  \vspace{-5pt}
  \small
  \setlength{\tabcolsep}{4pt}
  \begin{tabular}{@{}rrrc@{}}
    \toprule
    $\rho$ & Rank $\bar{r}$ & $\bar{r}/d_{\text{in}}$ & Coverage \\
    \midrule
    0.5   & 6     & 0.2\% & 0.544 \\
    0.7   & 20    & 0.6\% & 0.719 \\
    0.8   & 39    & 1.1\% & 0.810 \\
    0.9   & 106   & 3.0\% & 0.903 \\
    \textbf{0.95} & \textbf{250} & \textbf{7.0\%} & \textbf{0.948} \\
    0.99  & 928   & 25.9\% & 0.985 \\
    0.999 & 2,211 & 61.7\% & 0.997 \\
    \bottomrule
  \end{tabular}
\end{wraptable}
The energy fraction $\rho$ sets the rank retained by Eq.~\ref{eq:rank}, and Tab.~\ref{tab:rho} follows it across the spectrum of the stratified calibration set of 400 items, with $d_{\text{in}} = 3{,}584$ and the $\rho = 0.95$ row the projector used throughout. Coverage on held-out items tracks $\rho$ to within a point at every rank, so the estimate generalizes equally well whether it retains six directions or two thousand, and $\rho$ is a choice of how much of the input space to close. The rank grows slowly up to $\rho = 0.95$, where 95\% of the held-out energy is captured within 7\% of the dimensions, and steeply beyond it, with $\rho = 0.99$ retaining a quarter of the input space for four more points of coverage. We therefore use $0.95$, the last point at which the subspace is still small. Training under $0.90$, $0.95$, and $0.99$ in Tab.~\ref{tab:ablation} shows what the rank buys, as a larger $\rho$ suppresses more of the shortcut while the accuracy stays within seed variation.

\begin{table}[h]
  \centering
  \caption{The seven VideoLLMs, their language model family and parameter count, effective parameters for Gemma4-E2B, and the accuracy (\%) of the base model on the SSv2 evaluation split.}
  \label{tab:backbones}
  \small
  \begin{tabular}{@{}llrcc@{}}
    \toprule
    VideoLLM & Language model & Params & Full & Text-only \\
    \midrule
    Qwen2.5-VL-3B & Qwen2.5 & 3B & 38.5 & 22.4 \\
    Qwen2.5-VL-7B & Qwen2.5 & 7B & 49.1 & 19.3 \\
    Qwen2.5-VL-32B & Qwen2.5 & 32B & 53.1 & 24.9 \\
    Qwen3-VL-8B & Qwen3 & 8B & 51.5 & 30.0 \\
    LLaVA-OneVision-7B & Qwen2 & 7B & 39.2 & 24.8 \\
    Molmo2-O-7B & Olmo & 7B & 53.5 & 22.6 \\
    Gemma4-E2B & Gemma~4 & 2.3B & 25.8 & 25.7 \\
    \bottomrule
  \end{tabular}
\end{table}

\section{Implementation Details}
\label{app:impl}

\paragraph{Data splits} follow Sec.~\ref{sec:setup}. The blind-filtered SSv2 training set holds 56,922 items, of which 400 are drawn by stratified sampling over the families as the calibration set and held out, leaving 56,522 for training. The blind-filtered evaluation split holds 8,016 items, of which 200 drawn at random serve to select the checkpoint and the remaining 7,816 are the test set on which every accuracy of Sec.~\ref{sec:experiments} is reported. The calibration, training, selection, and test sets are therefore pairwise disjoint.

\paragraph{Backbones} of Sec.~\ref{sec:expsetup}, Qwen2.5-VL~\citep{bai2025qwen25vltechnicalreport}, Qwen3-VL~\citep{bai2025qwen3vltechnicalreport}, LLaVA-OneVision~\citep{li2025llavaonevision}, Molmo2~\citep{clark2026molmo2openweightsdata}, and Gemma~4~\citep{gemmateam2026gemma4technicalreport}, are listed in Tab.~\ref{tab:backbones} with their language model family and size, and with the accuracy of the base model on the SSv2 evaluation split with the full input and the text-only input.

\paragraph{Training} uses LoRA adapters~\citep{hu2021lora} on the attention and MLP projections of every decoder layer, with the down projection left untouched, and Tab.~\ref{tab:hparams} lists the settings. We optimize with AdamW~\citep{loshchilov2017decoupled} under a cosine schedule for up to 10 epochs with early stopping on the selection accuracy, where the Qwen2.5-VL-7B runs of SFT and VETO complete all 10 epochs for the epoch curve of Fig.~\ref{fig:epoch}(a), with the checkpoint selected as in every other run. Attention uses FlashAttention-2~\citep{dao2024flashattention} where the backbone supports it. Qwen2.5-VL-7B is trained with three seeds and reported as mean and standard deviation, the other backbones use the same three seeds, and Tab.~\ref{tab:seeds} gives the seed count of every row of Tab.~\ref{tab:compare} and Tab.~\ref{tab:ablation}. SFT and VETO share all of the above, and VETO adds only the projection of Sec.~\ref{sec:projection} after every optimizer step.

\paragraph{Calibration and projection} run once before training. For each adapted module the input activations at every text position of the 400 calibration prompts under the text-only input are collected from the base model, their uncentered second moment is decomposed, and the leading directions that carry $\rho = 0.95$ of the energy form $U_1^l$, which for Qwen2.5-VL-7B has rank 250 on average over the 168 adapted modules. The projection $A \leftarrow A\,(I - U_1^l {U_1^l}^\top)$ is applied to the $A$ factor of each adapter after every optimizer step. The two projection ablations of Tab.~\ref{tab:ablation} differ only in when this projection runs. At initialization only, the adapter is projected once and then trains without the projection, so the leakage grows from zero as the update re-enters the subspace. After training, the adapter of the SFT run is projected once. The AdamW moments are not projected in any run. What they accumulate along $U_1^l$ is removed from the weights at the next step, and the leakage $\|A^l U_1^l\|_F / \|A^l\|_F$ measured at every epoch stays at numerical zero in every VETO run, so the accumulation never reaches the weights. Since the projection is applied after every step, it is equivalent to discarding the component of each update inside $U_1^l$, so the constraint can equally be read as a hard regularizer on the update that sets its component along the shortcut directions to zero rather than penalizing it. Under the initialization-only ablation the same ratio rises from zero to $0.20$ after the first epoch and to $0.30$ by the fourth in each of the three seeds. Under SFT without any projection it stands at $0.34$ at the selected checkpoint in each of the three seeds, above the $0.26$ that a random direction of the same rank would give, so the update of SFT leans toward the text directions.

\begin{table}[t]
  \centering
  \caption{Hyperparameters shared by SFT and VETO.}
  \label{tab:hparams}
  \small
  \begin{tabular}{@{}ll@{}}
    \toprule
    LoRA rank / scaling / dropout & 16 / 32 / 0.05 \\
    Adapted modules & q, k, v, o, gate, up \\
    Optimizer & AdamW, lr $10^{-4}$, weight decay 0.01 \\
    Schedule & cosine, 10\% warm-up, clip 1.0 \\
    Items per update & 8 \\
    Precision & bf16 \\
    Epochs / patience & up to 10 / 3 on the selection set \\
    Frames / resolution & 16 / $\le 360 \times 420$ \\
    Micro-batch & 4 (Qwen, all ablation rows), 1 (others) \\
    Selection / test set & 200 / 7,816 items of the evaluation split \\
    Seeds & 3 per configuration unless stated \\
    Calibration & 400 prompts, text-only, $\rho = 0.95$ \\
    Hardware & one NVIDIA B200 per run \\
    \bottomrule
  \end{tabular}
  \vspace{-10pt}
\end{table}

\paragraph{Loss baselines} in Sec.~\ref{sec:ablation} follow the published objectives with the answer letter as the prediction target. Learned-Mixin +H~\citep{clark2019don} needs a bias model that carries the text-only pathway, for which we fine-tune the same backbone with the same LoRA on the text-only prompt and cache its letter logits $z_b$ for every training item, so that the bias model enters the loss as a fixed signal. The main model then trains on $z = \log\mathrm{softmax}(z_m) + g(h)\,\log\mathrm{softmax}(z_b)$, where $z_m$ are its own letter logits, $h$ its hidden state at the answer position, and $g(h) = \mathrm{softplus}(w^\top h + b)$ a learned gate. The input $h$ of the gate is normalized to unit RMS and the gate is clamped at 5. The loss is $\mathrm{CE}(z, y) + \lambda\,(-H(\mathrm{softmax}(g(h)\,\log\mathrm{softmax}(z_b))))$ at $\lambda = 0.36$, the value published for VQA-CP, and the model predicts from $z_m$ alone at evaluation. The LMH row trains under the budget of the other rows, up to 10 epochs with early stopping on the selection set. mDPO~\citep{wang2024mdpo} is applied as published to the instruction-tuned base model, which also serves as the reference policy, at learning rate $10^{-5}$ on the training items. The preferred response is the correct letter and the rejected one a wrong letter drawn per item, and the rejected video keeps a random fifth or less of the frames and blacks out the rest, following the crop of the original. The loss is the published sum of the preference, conditional, and anchored terms at $\beta = 0.1$ and $\delta = 0$.

\begin{wraptable}[6]{r}{0.26\linewidth}
  \vspace{-\baselineskip}
  \centering
  \caption{Accuracy under VCD, three seeds.}
  \label{tab:vcd}
  \vspace{-5pt}
  \small
  \setlength{\tabcolsep}{5pt}
  \begin{tabular}{@{}lcc@{}}
    \toprule
    & Full & VCD \\
    \midrule
    SFT  & 89.6{\tiny$\pm$0.4} & 89.7{\tiny$\pm$0.4} \\
    \textbf{VETO} & 89.2{\tiny$\pm$0.2} & 89.2{\tiny$\pm$0.2} \\
    \bottomrule
  \end{tabular}
\end{wraptable}
\paragraph{Decoding-time method.} Visual contrastive decoding (VCD)~\citep{leng2024mitigating} contrasts the letter logits under the original frames with those under noised frames at inference, and we implement it for our backbone with $\alpha = 1$, $\beta = 0.1$, and $T = 500$ as published. VCD is undefined without frames, so it leaves the text-only accuracy, and with it the shortcut, as it is. Tab.~\ref{tab:vcd} applies it to SFT and VETO on the SSv2 evaluation split, and neither the accuracy nor the shortcut moves under it. 

\paragraph{Training cost.} Tab.~\ref{tab:cost} measures the cost of each row of Tab.~\ref{tab:compare} under the main recipe. The one-off column is the calibration of VETO, the bias model and its cached logits of LMH, with the bias model extrapolated from a measured segment to its five epochs, and the cached reference log-probabilities of mDPO. Reshuffling costs as SFT, and the random subspace and the ablations cost as VETO. The per-update time is averaged over five runs and the one-off cost over three. VETO trains at nearly the cost of SFT, whereas LMH and mDPO cost several times more.

\begin{table}[t]
  \centering
  \caption{Training cost on one B200 for one epoch of the 57k items, mean $\pm$ standard deviation over repeated measurements.}
  \label{tab:cost}
  \small
  \begin{tabular}{@{}lcccc@{}}
    \toprule
    Method & One-off (GPU-h) & Per update (s) & One epoch (GPU-h) & $\times$SFT \\
    \midrule
    SFT & -- & 2.24$\pm$0.01 & 4.40$\pm$0.02 & 1.00 \\
    \textbf{VETO} & 0.03$\pm$0.00 & 2.36$\pm$0.02 & 4.67$\pm$0.05 & 1.06 \\
    LMH & 17.77$\pm$0.33 & 2.26$\pm$0.03 & 22.21$\pm$0.34 & 5.05 \\
    mDPO & 5.28$\pm$0.38 & 4.44$\pm$0.01 & 13.98$\pm$0.39 & 3.18 \\
    \bottomrule
  \end{tabular}
  \vspace{-10pt}
\end{table}

\begin{table}[h]
  \centering
  \caption{Accuracy and text-only accuracy (\%) on the SSv2 evaluation split for the base model and the rows of Tab.~\ref{tab:compare} and Tab.~\ref{tab:ablation}, Qwen2.5-VL-7B, mean $\pm$ standard deviation over $n$ seeds.}
  \label{tab:seeds}
  \small
  \begin{tabular}{@{}lccc@{}}
    \toprule
    & $n$ & Accuracy & Text-only accuracy \\
    \midrule
    Base & -- & 49.1 & 19.3 \\
    \midrule
    SFT & 3 & 89.6$\pm$0.4 & 49.8$\pm$0.8 \\
    Reshuffled options & 3 & 89.4$\pm$0.2 & 49.4$\pm$1.0 \\
    LMH & 1 & 83.4 & 32.6 \\
    mDPO & 1 & 87.0 & 45.3 \\
    Random subspace & 3 & 89.6$\pm$0.0 & 49.6$\pm$2.0 \\
    VETO ($\rho = 0.95$) & 3 & 89.2$\pm$0.2 & 25.3$\pm$0.6 \\
    \midrule
    $\rho = 0.90$ & 3 & 88.6$\pm$0.8 & 32.1$\pm$3.9 \\
    $\rho = 0.99$ & 3 & 89.5$\pm$0.1 & 20.4$\pm$0.4 \\
    Projected at init.\ only & 3 & 88.7$\pm$0.3 & 49.2$\pm$1.2 \\
    Projected after training & 3 & 82.4$\pm$0.4 & 21.6$\pm$0.1 \\
    \bottomrule
  \end{tabular}
  \vspace{-10pt}
\end{table}

\paragraph{Seed statistics for Sec.~\ref{sec:ablation}.} Tab.~\ref{tab:seeds} lists, for the base model and every row of Tab.~\ref{tab:compare} and Tab.~\ref{tab:ablation}, the mean and standard deviation over seeds of the accuracy and the text-only accuracy on the SSv2 evaluation split, from which Text $\Delta$ is the mean text-only accuracy minus that of the base model.

\begin{table}[t]
  \centering
  \caption{Fraction of text-position activation energy inside $U_1^l$ on the base Qwen2.5-VL-7B, averaged over modules and by module type, under the text-only input on the calibration items, on held-out SSv2, and on the six benchmarks under shift, and under the original video on the calibration items.
  }
  \label{tab:coverage_cond}
  \small
  \begin{tabular}{@{}lccccccc@{}}
    \toprule
    Input & Mean & q & k & v & o & gate & up \\
    \midrule
    Text-only, calibration items & 0.950 & 0.950 & 0.950 & 0.950 & 0.950 & 0.950 & 0.950 \\
    Text-only, held-out SSv2 & 0.947 & 0.948 & 0.948 & 0.948 & 0.945 & 0.946 & 0.946 \\
    Text-only, Perception Test & 0.806 & 0.818 & 0.818 & 0.818 & 0.789 & 0.797 & 0.797 \\
    Text-only, Vinoground & 0.823 & 0.832 & 0.832 & 0.832 & 0.807 & 0.818 & 0.818 \\
    Text-only, AoTBench & 0.778 & 0.791 & 0.791 & 0.791 & 0.756 & 0.771 & 0.771 \\
    Text-only, MVBench & 0.808 & 0.820 & 0.820 & 0.820 & 0.790 & 0.800 & 0.800 \\
    Text-only, TVBench & 0.786 & 0.800 & 0.800 & 0.800 & 0.757 & 0.781 & 0.781 \\
    Text-only, TempCompass & 0.839 & 0.850 & 0.850 & 0.850 & 0.826 & 0.830 & 0.830 \\
    Original video, calibration items & 0.879 & 0.891 & 0.891 & 0.891 & 0.851 & 0.876 & 0.876 \\
    \bottomrule
  \end{tabular}
  \vspace{-10pt}
\end{table}

\section{The Two Pathways in Detail}
\label{app:sec44}

\paragraph{Where the video enters the text positions.} The subspace $U_1^l$ is estimated from text-position activations under the text-only input, and the constraint keeps the update out of it alone. Whether the video can still reach the update therefore depends on where the text-position activations fall once video tokens are present. We measure, on the base model and the calibration items, the fraction of text-position activation energy that lies inside $U_1^l$, that is, $\mathrm{tr}((U_1^l)^\top C^l U_1^l)/\mathrm{tr}(C^l)$ with $C^l$ the uncentered second moment of the activations at the positions after the last vision token, under the text-only input and under the original video. Under the text-only input the fraction equals the $\rho$ imposed by construction. Under the original video it falls, as Tab.~\ref{tab:coverage_cond} shows, most at the output projection of attention, which is where the attention output carries what was read from the video tokens. The video thus moves part of the text-position activations outside the calibrated subspace, and this part is where the update remains free, which is the route through which VETO learns from the video. The same table gives the fraction under the text-only input on held-out SSv2 items and on the text of the six benchmarks under shift. On held-out SSv2 it stays at the imposed value, and on the six benchmarks, whose text shares nothing with SSv2, it is close to four fifths, against about one fourteenth for a random subspace of the same rank, so the subspace estimated on the text of SSv2 covers most of where the text alone is read on other text as well. The calibration collects these activations without video tokens in front of the text, so the text tokens sit at the positions of the text-only input. The subspace nevertheless holds close to nine tenths of the text-position energy once the video tokens are present, as the table shows, so the shift of positions between the two conditions does not move the text activations out of it. A fifth of the energy of the text-only activations of the six benchmarks lies outside the subspace, yet the text-only gain on them stays well below that of SFT, so the directions the shortcut uses lie mostly inside the calibrated subspace even for text the calibration never saw.

\paragraph{Entropy over the options.} The entropy reported in Sec.~\ref{sec:redistribution} is that of the distribution the VideoLLM places over the option letters at the answer position, $-\sum_{o} p(o)\log p(o)$ in nats, averaged over the evaluation items, so that it runs from zero, where one option takes all the probability, to $\ln n$ for $n$ options, where the options are equally likely. The share of the maximum in Tab.~\ref{tab:sec44ent} divides the entropy of each item by the $\ln n$ of its own options. The last column is the difference between the first two, that is, the part of the entropy without the video that the video resolves. Tab.~\ref{tab:entropy_raw} gives the values in nats, averaged over three seeds on 2,000 items of each set.

\begin{table}[h]
  \centering
  \caption{Entropy over the options in nats, mean $\pm$ sd over three seeds, on 2,000 items of the SSv2 evaluation split and of Perception Test, Qwen2.5-VL-7B.}
  \label{tab:entropy_raw}
  \small
  \begin{tabular}{@{}lcccc@{}}
    \toprule
    & \multicolumn{2}{c}{SSv2} & \multicolumn{2}{c}{Perception Test} \\
    \cmidrule(lr){2-3}\cmidrule(lr){4-5}
    & Text-only & Full & Text-only & Full \\
    \midrule
    Maximum ($\ln 4$ or $\ln 3$) & 1.39 & 1.39 & 1.10 & 1.10 \\
    Base & 1.04 & 0.84 & 0.90 & 0.63 \\
    SFT & 0.22{\tiny$\pm$0.08} & 0.03{\tiny$\pm$0.01} & 0.37{\tiny$\pm$0.09} & 0.16{\tiny$\pm$0.05} \\
    \textbf{VETO} & 0.93{\tiny$\pm$0.08} & 0.03{\tiny$\pm$0.02} & 0.89{\tiny$\pm$0.02} & 0.20{\tiny$\pm$0.07} \\
    \bottomrule
  \end{tabular}
  \vspace{-5pt}
\end{table}

\paragraph{The answer through the layers.} Reading the option logits from the hidden state of every layer through the final norm and the output head~\citep{nostalgebraist2020logitlens} gives the accuracy each layer would reach on its own. Under the text-only input the base model and VETO stay at chance at every layer, whereas SFT rises above chance from the last third of the layers and reaches its text-only accuracy at the end, so the text-only pathway forms in the same late layers where Fig.~\ref{fig:sec44a} shows the representation drifting most. With the video, SFT and VETO rise at the same layers to nearly the same accuracy, so the constraint does not change where the answer with the video is formed. Tab.~\ref{tab:lens} gives the accuracy at a quarter, a half, three quarters, and the full depth.

\begin{table}[h]
  \centering
  \caption{Accuracy read from the hidden state at four depths, under the text-only input and with the original video, on 2,000 items of the SSv2 evaluation split, Qwen2.5-VL-7B, mean over three seeds.}
  \label{tab:lens}
  \small
  \begin{tabular}{@{}llcccc@{}}
    \toprule
    Input & Model & $L/4$ & $L/2$ & $3L/4$ & $L$ \\
    \midrule
    \multirow{3}{*}{Text-only} & Base & 0.26 & 0.27 & 0.23 & 0.20 \\
                               & SFT  & 0.26 & 0.26 & 0.40 & 0.50 \\
                               & VETO & 0.27 & 0.27 & 0.27 & 0.25 \\
    \midrule
    \multirow{3}{*}{Original video} & Base & 0.25 & 0.26 & 0.29 & 0.49 \\
                                    & SFT  & 0.25 & 0.26 & 0.52 & 0.89 \\
                                    & VETO & 0.27 & 0.26 & 0.47 & 0.89 \\
    \bottomrule
  \end{tabular}
  \vspace{-5pt}
\end{table}

\paragraph{The video still enters the decision.} Blocking the attention from text tokens to video tokens at inference~\citep{geva2023dissecting} leaves the input intact and removes only the reading of it, where the text-only input omits the video itself, and Tab.~\ref{tab:reliance_full} reports both. On SSv2 SFT loses most of its accuracy under the block, so the video enters its decision, and what remains is its accuracy from the text alone, so the shortcut keeps answering when the video cannot be read. The base model retains far less under the block, and this gap is what fine-tuning added to the shortcut. On Perception Test the block costs the base model more than SFT, so the video there carries evidence and SFT reads less of it than the base model, while what SFT retains is again its accuracy from the text alone. The shortcut therefore sits beside the reading of the video on the training distribution and in place of part of it under the shift, and in neither case is the video path closed.

\paragraph{Reliance on the video.} The last rows of Tab.~\ref{tab:reliance_full} give the share of the answers each model gets right with the full input that it gets right under the block. In every cell the accuracy under the block stays close to the accuracy from text alone, within a point under SFT, the base model, and the random subspace and within five points under VETO, so the block removes the reading of the video nearly as cleanly as removing the video itself. A random subspace of the same rank at every module leaves the accuracy under the block under the shift at the level of SFT and the accuracy within a point of it, inside the seed spread, so the gain VETO makes there rests on the directions its calibration locates rather than on the rank it removes.

\begin{table}[t]
  \centering
  \caption{Accuracy (\%) with the full input, with the attention to the video blocked, and from the text alone, on SSv2 and Perception Test, Qwen2.5-VL-7B, mean $\pm$ sd over three seeds.
  }
  \label{tab:reliance_full}
  \small
  \setlength{\tabcolsep}{3pt}
  \begin{tabular}{@{}lccccccc@{}}
    \toprule
    & \multicolumn{3}{c}{SSv2} & & \multicolumn{3}{c}{Perception Test} \\
    \cmidrule{2-4}\cmidrule{6-8}
    & Full & Blocked & Text-only & & Full & Blocked & Text-only \\
    \midrule
    Base & 49.1 & 19.5 & 19.3 & & 61.1 & 21.5 & 21.3 \\
    SFT  & 89.6{\tiny$\pm$0.4} & 49.5{\tiny$\pm$1.1} & 49.8{\tiny$\pm$0.8} & & 61.1{\tiny$\pm$0.7} & 38.4{\tiny$\pm$0.9} & 38.3{\tiny$\pm$1.1} \\
    Random subspace & 89.6{\tiny$\pm$0.0} & 49.8{\tiny$\pm$1.8} & 49.6{\tiny$\pm$2.0} & & 62.1{\tiny$\pm$0.9} & 35.8{\tiny$\pm$0.5} & 35.2{\tiny$\pm$0.6} \\
    \textbf{VETO} & 89.2{\tiny$\pm$0.2} & 30.2{\tiny$\pm$1.2} & 25.3{\tiny$\pm$0.6} & & 64.3{\tiny$\pm$0.7} & 25.9{\tiny$\pm$3.1} & 26.2{\tiny$\pm$2.2} \\
    \midrule
    \multicolumn{8}{@{}l}{\textit{Share of correct answers kept under the block}} \\
    Base & & 29.7 & & & & 25.2 & \\
    SFT  & & 51.2{\tiny$\pm$1.1} & & & & 45.7{\tiny$\pm$2.7} & \\
    \textbf{VETO} & & 30.2{\tiny$\pm$1.1} & & & & 26.3{\tiny$\pm$4.1} & \\
    \bottomrule
  \end{tabular}
\end{table}

\section{Fine-Tuning on Other Datasets}
\label{app:dataset_choice}

\begin{wraptable}[11]{r}{0.42\linewidth}
  \vspace{-\baselineskip}
  \centering
  \caption{Fine-tuning on NExT-QA and STAR, mean $\pm$ sd over three seeds.}
  \label{tab:other_data}
  \vspace{-5pt}
  \small
  \setlength{\tabcolsep}{4pt}
  \begin{tabular}{@{}llcc@{}}
    \toprule
    Dataset & Model & Acc. & Text $\Delta$ \\
    \midrule
    \multirow{3}{*}{NExT-QA} & Base & 70.9 & -- \\
                             & SFT & 71.7{\tiny$\pm$0.4} & +8.1{\tiny$\pm$0.7} \\
                             & \textbf{VETO} & 71.6{\tiny$\pm$0.2} & +1.3{\tiny$\pm$0.2} \\
    \midrule
    \multirow{3}{*}{STAR} & Base & 67.9 & -- \\
                          & SFT & 70.5{\tiny$\pm$0.2} & +11.6{\tiny$\pm$1.1} \\
                          & \textbf{VETO} & 70.9{\tiny$\pm$0.6} & +1.3{\tiny$\pm$0.4} \\
    \bottomrule
  \end{tabular}
\end{wraptable}
The main experiments fine-tune on SSv2 alone. This section reports fine-tuning on two further datasets, NExT-QA~\citep{xiao2021next} and STAR~\citep{wu2021star}, whose questions and options are written freely rather than from templates. Tab.~\ref{tab:other_data} reports fine-tuning Qwen2.5-VL-7B under the recipe of Sec.~\ref{sec:expsetup}, on the blind-filtered training split of NExT-QA and a 20k subsample of that of STAR stratified over its question types, with the blind filter and calibration applied as to SSv2 in the main experiments. On both datasets SFT raises text-only accuracy far more than VETO, whereas the gain on the full input is nearly the same under the two, as across the backbones of Fig.~\ref{fig:veto}.

\end{document}